\documentclass{article}

\usepackage[preprint]{neurips_2019}

\usepackage[utf8]{inputenc} 
\usepackage[T1]{fontenc}    
\usepackage{url}            
\usepackage{booktabs}       
\usepackage{amsfonts}       
\usepackage{nicefrac}       
\usepackage{microtype}      
\usepackage{amsmath}
\usepackage{graphicx}
\usepackage{algorithm}
\usepackage{algpseudocode}
\usepackage{bbm}
\usepackage{mathtools}
\usepackage{caption}
\usepackage[subrefformat=parens]{subcaption}

\title{HyperGCN: A New Method of Training Graph Convolutional Networks on Hypergraphs}

\author{%
  Naganand Yadati \\
  Indian Insitute of Science, Bangalore\\
  \texttt{y.naganand@gmail.com}
  \And
  Madhav Nimishakavi \\
  Indian Insitute of Science, Bangalore\\
  \And
  Prateek Yadav \\
  Indian Insitute of Science, Bangalore\\
  \And
  Vikram Nitin \\
  Indian Insitute of Science, Bangalore\\
  \And
   Anand Louis \\
   Indian Insitute of Science, Bangalore\\
   \And
   Partha Talukdar \\
   Indian Insitute of Science, Bangalore\\  
}

\newtheorem{definition}{Definition}
\DeclareMathOperator*{\argmax}{arg\,max}
\newcommand{\method}[1]{HyperGCN}

\begin{document}
\maketitle

\begin{abstract}
	In many real-world network datasets such as co-authorship, co-citation, email communication, etc., relationships are complex and go beyond pairwise associations.
	Hypergraphs provide  a flexible and natural modeling tool to model such complex relationships.
	The obvious existence of such complex relationships in many real-world networks naturally motivates the problem of learning with hypergraphs.
	A popular learning paradigm is hypergraph-based semi-supervised learning (SSL) where the goal is to assign labels to initially unlabelled vertices in a hypergraph.
	Motivated by the fact that a graph convolutional network (GCN) has been effective for graph-based SSL, we propose \method{}, a novel way of training a GCN for SSL on hypergraphs.
	We demonstrate \method{}'s effectiveness through detailed experimentation on real-world hypergraphs and analyse when it is going to be more effective than state-of-the art baselines.
\end{abstract}
\section{Introduction}

In many real-world network datasets such as co-authorship, co-citation, email communication, etc., relationships are complex and go beyond pairwise associations.
Hypergraphs provide  a flexible and natural modeling tool to model such complex relationships.
For example, in a co-authorship network an author (hyperedge) can be a co-author of more than two documents (vertices).

The obvious existence of such complex relationships in many real-world networks naturaly motivates the problem of learning with hypergraphs \cite{lhg06,sslhg13,sslhg17,hgnn_aaai19}.
A popular learning paradigm is graph-based / hypergraph-based semi-supervised learning (SSL) where the goal is to assign labels to initially unlabelled vertices in a graph / hypergraph \cite{csz_ssl10,sslintro09,ssl_ppt_14}.
While many techniques have used explicit Laplacian regularisation in the objective \cite{sslintronips03,sslintroicml03,ca02,dlvssl_icml08}, the state-of-the-art neural methods encode the graph / hypergraph structure $G=(V,E)$ \textit{implicitly} via a neural network  $f(G,X)$\cite{gcniclr17,dcnn_nips16,hgnn_aaai19} ($X$ contains the initial features on the vertices for example, text attributes for documents).

While explicit Laplacian regularisation assumes similarity among vertices in each edge / hyperedge, implicit regularisation of graph convolutional networks (GCNs) \cite{gcniclr17} avoids this restriction and enables application to a broader range of problems in combinatorial optimisation \cite{gatmco_kdd19,gcgnn_ijcai19,tspgnn_aaai19,gtsgcn_nips18}, computer vision \cite{mlgcn_cvpr19,vqagcn_nips18,zero_shot_gcn_cvpr18}, natural language processing \cite{wordgcn_acl19,textgcn_aaai19,gcn_srl_emnlp17}, etc.
In this work, we propose, \method{}, a novel training scheme for a GCN on hypergraph and show its effectiveness not only in SSL where hyperedges encode similarity but also in combinatorial optimisation where hyperedges do not encode similarity.
Combinatorial optimisation on hypergraphs has recently been highlighted as crucial for real-world network analysis \cite{hittingset_arxiv19,kcover_arxiv19}.

Methodologically, \method{} approximates each hyperedge of the hypergraph by a set of pairwise edges connecting the vertices of the hyperedge and treats the learning problem as a graph learning problem on the approximation.
While the state-of-the-art hypergraph neural networks (HGNN) \cite{hgnn_aaai19} approximates each hyperedge by a clique and hence requires $\ ^sC_2$ (quadratic number of) edges for each hyperedge of size $s$, our method, i.e. \method{}, requires a linear number of edges (i.e. $O(s)$) for each hyperedge.
The advantage of this linear approximation is evident in Table \ref{time_table} where a faster variant of our method has lower training time on synthetic data (with higher density as well) for densest $k$-subhypergraph and SSL on real-world hypergraphs (DBLP and Pubmed).
 In summary, we make the following contributions:

\begin{table}[t]
	\centering
	{
		\begin{tabular}{ |c|cc|c|c|  }
			\hline
		Model$\downarrow$\quad Metric $\rightarrow$ & Training time & Density & DBLP & Pubmed\\
			\hline
			HGNN & $170$s & $337$ & $0.115$s & $0.019$s\\
			\hline
			Fast\method{} & $\mathbf{143}$s & $\mathbf{352}$ & $\mathbf{0.035}$s & $\mathbf{0.016}$s\\
			\hline
	\end{tabular}}
	\caption{\label{time_table} average training time of an epoch (lower is better)} 
\end{table}

 \begin{itemize}
 	\item We propose \method{}, a novel method of training a graph convolutional network (GCN) on hypergraphs using existing tools from spectral theory of hypergraphs (Section \ref{sec:method}).
 	\item We apply \method{} to the problems of SSL on attributed hypergraphs and combinatorial optimisation. 
 	Through detailed experimentation, we demonstrate its effectiveness compared to the state-of-the art HGNN  \cite{hgnn_aaai19} and other baselines (Sections \ref{sec:rel_exp}, and \ref{sec:co_exp}). 
 	\item We thoroughly discuss when we prefer \method{} to HGNN (Sections \ref{sec:analysis}, and \ref{train_time})
 \end{itemize}

 While the motivation of \method{} is based on similarity of vertices in a hyperedge, we show that it can be used effectively for combinatorial optimisation where hyperedges do not encode similarity.

\section{Related work}
\label{sec:related_work}

In this section, we discuss related work and then the background in the next section.

\textbf{Deep learning on graphs}: {\it Geometric deep learning} \cite{gdl17} is an umbrella phrase for emerging techniques attempting to generalise (structured) deep neural network models to non-Euclidean domains such as graphs and manifolds. 
Graph convolutional network (GCN) \cite{gcniclr17} defines the convolution using a simple linear function of the graph Laplacian and is shown to be effective on semi-supervised classification on attributed graphs.
The reader is referred to a comprehensive literature review \cite{gdl17} and extensive surveys \cite{grl17,gnet_arxiv18,dlgsurvey_arxiv18,aagsurvey_arxiv18,gnnsurvey_arxiv19} on this topic of deep learning on graphs.

\textbf{Learning on hypergraphs}: The clique expansion of a hypergraph was introduced in a seminal work \cite{lhg06} and has become popular \cite{holg06,hg_ijcai15,poh18}. 
Hypergraph neural networks \cite{hgnn_aaai19} use the clique expansion to extend GCNs for hypergraphs.
Another line of work uses mathematically appealing tensor methods \cite{teeccv06,gthg09,td09}, but they are limited to uniform hypergraphs. 
Recent developments, however, work for arbitrary hypergraphs and fully exploit the hypergraph structure \cite{sslhg13,sslhg17,laplacian_mediators_cocoon18,shg_icml18,hgal_aistats19}. 

\textbf{Graph-based SSL}: 
Researchers have shown that using unlabelled data in training can improve learning accuracy significantly.
This topic is so popular that it has influential books \cite{csz_ssl10,sslintro09,ssl_ppt_14}.

\textbf{Graph neural networks for combinatorial optimisation}:
Graph-based deep models have recently been shown to be effective as learning-based approaches for NP-hard problems such as maximal independent set, minimum vertex cover, etc. \cite{gtsgcn_nips18}, the decision version of the traveling salesman problem \cite{tspgnn_aaai19}, graph colouring \cite{gcgnn_ijcai19}, and clique optimisation \cite{gatmco_kdd19}.
\section{Background: Graph convolutional network}
\label{sec:background}

Let $\mathcal{G} = (\mathcal{V}, \mathcal{E})$, with $N = |\mathcal{V}|$, be a simple undirected graph with adjacency $A\in\mathbb{R}^{N\times N}$, and data matrix $X\in\mathbb{R}^{N\times p}$.
which has $p$-dimensional real-valued vector representations for each node $v\in\mathcal{V}$. 

The basic formulation of graph convolution \cite{gcniclr17} stems from the convolution theorem \cite{ctmallat99} and it can be shown that the convolution $C$ of a real-valued graph signal $S\in\mathbb{R}^N$ and a filter signal $F\in\mathbb{R}^N$ is approximately $C \approx (w_0 + w_1\tilde{L})S$ where $w_0$ and $w_1$ are learned weights, and $\tilde{L} = \frac{2L}{\lambda_N} - I$ is the scaled graph Laplacian, $\lambda_N$ is the largest eigenvalue of the symmetrically-normalised graph Laplacian $L = I - D^{-\frac{1}{2}}AD^{-\frac{1}{2}}$ where $D = \text{diag}(d_1,\cdots,d_N)$ is the diagonal degree matrix with elements $d_i=\sum_{j=1,j\neq i}^NA_{ji}$.
The filter $F$ depends on the structure of the graph (the graph Laplacian $L$).
The detailed derivation from the convolution theorem uses existing tools from graph signal processing \cite{shuman13,wgsgt11,gdl17} and is provided in the supplementary material.
The key point here is that the convolution of two graph signals is a {\it linear function} of the graph Laplacian $L$.
 \begin{table*}[!htbp]
	\caption{Summary of symbols used in the paper.}
	\label{notation_table}
	\centering
	\small
	\begin{tabular}{ll|ll}
		\hline
		Symbol     & Description & Symbol  & Description      \\
		\hline
		$\mathcal{G} = (\mathcal{V}, \mathcal{E})$ & an undirected simple graph   & $\mathcal{H} = (V,E)$ & an undirected hypergraph \\
		$\mathcal{V}$     & set of nodes  & $V$  & set of hypernodes  \\
		$\mathcal{E}$     & set of edges    & $E$  & set of hyperedges     \\
		$N=|\mathcal{V}|$     & number of nodes    & $n=|V|$  & number of  hypernodes    \\
		$L$     & graph Laplacian    & $\mathbb{L}$  & hypergraph Laplacian \\
		$A$     & graph adjacency matrix   & $H$  & hypergraph incidence matrix \\
		\hline
	\end{tabular}
\end{table*}

The graph convolution for $p$ different graph signals contained in the data matrix $X\in\mathbb{R}^{N\times p}$ with learned weights $\Theta\in\mathbb{R}^{p\times r}$ with $r$ hidden units is $\bar{A}X\Theta$
$,\quad \bar{A} = \tilde{D}^{-\frac{1}{2}}\tilde{A}\tilde{D}^{-\frac{1}{2}},\ \tilde{A} = A + I,\ \text{and } \tilde{D}_{ii} = \sum_{j=1}^N\tilde{A}_{ij}$.
The proof involves a renormalisation trick \cite{gcniclr17} and is in the supplementary.

\paragraph{GCN \cite{gcniclr17}} 
The forward model for a simple two-layer GCN takes the following simple form:
\begin{equation}
\label{gcn}
Z = f_{GCN}(X,A) = \text{softmax}\Bigg(\bar{A}\ \ \text{ReLU}\bigg(\bar{A}X\Theta^{(1)}\bigg)\Theta^{(2)}\Bigg),
\end{equation}
where $\Theta^{(1)}\in\mathbb{R}^{p\times h}$ is an input-to-hidden weight matrix for a hidden layer with $h$ hidden units and $\Theta^{(2)}\in\mathbb{R}^{h\times r}$ is a  hidden-to-output weight matrix.
The softmax activation function is defined as $\text{softmax}(x_i) = \frac{\text{exp}(x_i)}{\sum_j\text{exp}(x_j)}$ and applied row-wise.

\textbf{GCN training for SSL}: For multi-class, classification with $q$ classes, we minimise cross-entropy,
\begin{equation}
\label{sslgcnloss}
\mathcal{L} = -\sum_{i\in \mathcal{V}_L}\sum_{j=1}^q Y_{ij}\ln Z_{ij},
\end{equation}
over the set of labelled examples $\mathcal{V}_L$. Weights $\Theta^{(1)}$ and $\Theta^{(2)}$ are trained using gradient descent. 

A summary of the notations used throughout our work is shown in Table \ref{notation_table}.
\section{\method{}: Hypergraph Convolutional Network}
\label{sec:method}

We consider semi-supervised hypernode classification on an undirected hypergraph $\mathcal{H} = (V,E)$ with $|V|=n$, $|E|=m$ and a small set $V_L$ of labelled hypernodes.
Each hypernode $v\in V = \{1,\cdots, n\}$ is also associated with a feature vector $x_v\in\mathbb{R}^p$ of dimension $p$ given by $X\in\mathbb{R}^{n\times p}$.
The task is to predict the labels of all the unlabelled hypernodes, that is, all the hypernodes in the set $V \setminus V_L$. 

\textbf{Overview}: The crucial working principle here is that the hypernodes in the same hyperedge are similar and hence are likely to share the same label \cite{sslhg17}.
Suppose we use $\{h_v : v \in V\}$ to denote some representation of the hypernodes in $V$, then, for any $e \in E$, the function $\max_{i,j \in e}||h_i - h_j||^2$ will be ``small'' only if vectors corresponding to the hypernodes in $e$ are ``close'' to each other. 
Therefore, $\sum_{e \in E} \max_{i,j \in e}||h_i - h_j||^2$ as a regulariser is likely to achieve the objective of the hypernodes in the same hyperedge having similar representations. 
However, instead of using it as an explicit regulariser, we can achieve the same goal by using GCN over an appropriately defined Laplacian of the hypergraph. 
In other words, we use the notion of {\em hypergraph Laplacian} as an implicit regulariser which achieves this objective.

\begin{figure*}
	\begin{center}
		\includegraphics[width=\textwidth,height=\textheight,keepaspectratio]{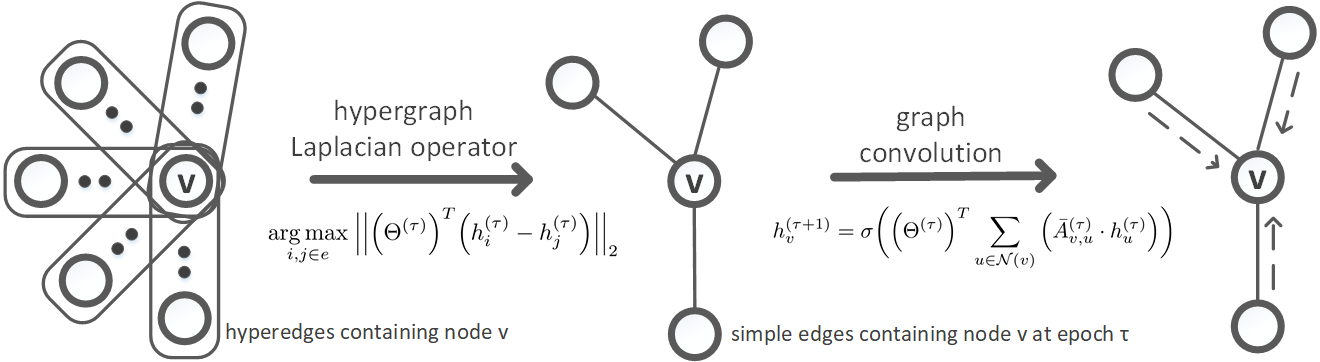}
		\caption{\label{fig:fig0}Graph convolution on a hypernode $v$ using \method{}.}
	\end{center}  
\end{figure*}

A hypergraph Laplacian with the same underlying motivation as stated above was proposed in prior works \cite{hgl_18,hg_stoc15}. We present this Laplacian first. Then we run GCN over the simple graph associated with this hypergraph Laplacian. We call the resulting method $1$-\method{} (as each hyperedge is approximated by exactly one pairwise edge). One epoch of $1$-\method{} is  shown in figure \ref{fig:fig0}

\subsection{Hypergraph Laplacian}
\label{sec:hyper_laplacian}

 As explained before, the key element for a GCN is the graph Laplacian of the given graph $\mathcal{G}$. 
 Thus, in order to develop a GCN-based SSL method for hypergraphs, we first need to define a Laplacian for hypergraphs.
One such way \cite{hgl_18} (see also \cite{hg_stoc15}) is a non-linear function $\mathbb{L} : \mathbb{R}^n \rightarrow \mathbb{R}^n$ (the Laplacian matrix for graphs can be viewed as a linear function $L : \mathbb{R}^n \rightarrow \mathbb{R}^n$).

\begin{definition}[Hypergraph Laplacian \cite{hgl_18,hg_stoc15}\footnotemark]
\label{def:hyp_lap}
Given a real-valued signal $S \in \mathbb{R}^{n}$ defined on the hypernodes, $\mathbb{L}(S)$ is computed as follows.
 \begin{enumerate}
\item For each hyperedge $e \in E$, let $(i_e, j_e) \coloneqq \text{argmax}_{i,j \in e} |S_i - S_j|$, breaking ties randomly\footnotemark[\value{footnote}].
\footnotetext{The problem of breaking ties in choosing $i_e$ (resp. $j_e$) is a non-trivial problem as shown in \cite{hgl_18}. Breaking ties randomly was proposed in \cite{hg_stoc15}, but \cite{hgl_18} showed that this might not work for all applications (see \cite{hgl_18} for more details).
\cite{hgl_18} gave a way to break ties, and gave a proof of correctness for their tie-breaking rule for the problems they studied. We chose to break ties randomly because of its simplicity and its efficiency. }	
  \item A weighted graph $G_S$ on the vertex set $V$ is constructed by adding edges $\{\{i_e, j_e\} : e \in E\}$ with weights $w(\{i_e, j_e\}) \coloneqq w(e)$ to $G_S$, where $w(e)$ is the weight of the hyperedge $e$. Next, to each vertex $v$, self-loops are added such that the degree of the vertex in $G_S$ is equal to $d_v$. Let $A_S$ denote the weighted adjacency matrix of the graph $G_S$.
  \item The symmetrically normalised hypergraph Laplacian is ${\mathbb{L}}(S) \coloneqq (I - D^{-\frac{1}{2}} A_S D^{-\frac{1}{2}}) S$

 \end{enumerate}
\end{definition}


\subsection{$1$-HyperGCN}
\label{sec:hypergcn}


By following the Laplacian construction steps outlined in Section \ref{sec:hyper_laplacian}, we end up with the simple graph $G_S$ with normalized adjacency matrix $\bar{A}_S$. We now perform GCN over this simple graph $G_S$. The graph convolution operation in Equation \eqref{gcn}, when applied to a hypernode $v \in V$ in $G_S$, in the neural message-passing framework  \cite{mpnn_icml17} is
$h_{v}^{(\tau + 1)} = \sigma\bigg((\Theta^{(\tau)})^T\sum_{u \in \mathcal{N}(v)}([\bar{A}^{(\tau)}_S]_{v,u} \cdot h^{(\tau)}_{u})\bigg)$.
Here, $\tau$ is epoch number, $h^{(\tau + 1)}_{v}$ is the new hidden layer representation of node $v$, $\sigma$ is a non-linear activation function, $\Theta$ is a matrix of learned weights, $\mathcal{N}(u)$ is the set of neighbours of $v$, $[\bar{A}^{(\tau)}_S]_{v,u}$ is the weight on the edge $\{v,u\}$ after normalisation, and $h^{(\tau)}_{u}$ is the previous hidden layer representation of the neighbour $u$. We note that along with the embeddings of the hypernodes, the adjacency matrix is also re-estimated in each epoch.

Figure \ref{fig:fig0} shows a hypernode $v$ with five hyperedges incident on it.
We consider exactly one representative simple edge for each hyperedge $e\in E$ given by $(i_e, j_e)$ where $(i_e, j_e) = \argmax_{i,j\in e}||(\Theta^{(\tau)})^T(h^{(\tau)}_i-h^{(\tau)}_j)||_2$ for epoch $\tau$.
Because of this consideration, the hypernode $v$ may not be a part of all representative simple edges (only three shown in figure).
We then use traditional Graph Convolution Operation on $v$ considering only the simple edges incident on it.
Note that we apply the operation on each hypernode $v\in V$ in each epoch $\tau$ of training until convergence.

\textbf{Connection to total variation on hypergraphs}: Our 1-\method{} model can be seen as performing implicit 
regularisation based on the total variation on hypergraphs \cite{sslhg13}. 
In that prior work, explicit regularisation and only the hypergraph structure is used for hypernode classification in the SSL setting. 
\method{}, on the other hand, can use both the hypergraph structure and also exploit any available features on the hypernodes, e.g., text attributes for documents.

\begin{figure*}
	\label{fig1}
	\begin{center}
		\includegraphics[width=\textwidth,height=\textheight,keepaspectratio]{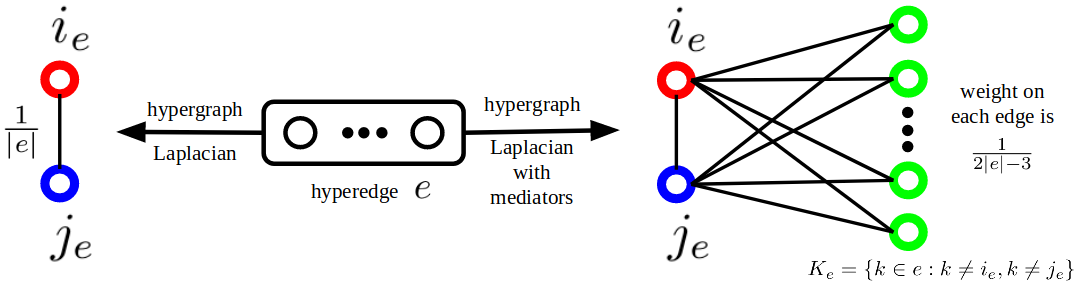}
		\caption{\label{fig:hmlap}Hypergraph Laplacian \cite{hgl_18} vs. the generalised hypergraph Laplacian with mediators \cite{laplacian_mediators_cocoon18}. Our approach requires at most a linear number of edges ($1$ and $2|e|-3$ respectively) while HGNN \cite{hgnn_aaai19} requires a quadratic number of edges for each hyperedge.}
	\end{center}  
\end{figure*}

\subsection{\method{}: Enhancing $1$-HyperGCN with mediators}
\label{sec:mediators}

One peculiar aspect of the hypergraph Laplacian discussed is that each hyperedge $e$ is represented by a single pairwise simple edge $\{i_e, j_e\}$ (with this simple edge potentially changing from epoch to epoch).
This hypergraph Laplacian ignores the hypernodes in $K_e:=\{k\in e:\ k\neq i_e, k\neq j_e\}$ in the given epoch.
Recently, it has been shown that a generalised hypergraph Laplacian in which the hypernodes in $K_e$ act as ``mediators" \cite{laplacian_mediators_cocoon18} satisfies all the properties satisfied by the above Laplacian given by \cite{hgl_18}.
The two Laplacians are pictorially compared in Figure \ref{fig:hmlap}.
Note that if the hyperedge is of size $2$, we connect $i_e$ and $j_e$ with an edge.
We also run a GCN on the simple graph associated with the hypergraph Laplacian with mediators \cite{laplacian_mediators_cocoon18} (right in Figure \ref{fig:hmlap}).
It has been suggested that the weights on the edges for each hyperedge in the hypergraph Laplacian (with mediators) sum to $1$ \cite{laplacian_mediators_cocoon18}.
We chose each weight to be $\frac{1}{2|e|-3}$ as there are $2|e|-3$ edges for a hyperedge $e$.

\subsection{Fast\method{}}

We use just the initial features $X$ (without the weights) to construct the hypergraph Laplacian matrix (with mediators) and we call this method Fast\method{}.
Because the matrix is computed only once before training (and not in each epoch), the training time of Fast\method{} is much less than that of other methods.
We have provided the algorithms for the three methods in the supplementary.
\section{Experiments for semi-supervised learning}
\label{sec:rel_exp}
\setlength{\arrayrulewidth}{1pt}

We conducted experiments not only on real-world datasets but also on categorical data (results in supplementary) which are a standard practice in hypergraph-based learning \cite{lhg06,sslhg13,sslhg17,shg_icml18,dsfm_neurips18,qdsfm_neurips18}.

\subsection{Baselines}

We compared \method{}, $1$-\method{} and Fast\method{} against the following baselines:
  \begin{itemize}
    \item \textbf{Hypergraph neural networks (HGNN) \cite{hgnn_aaai19}} uses the clique expansion \cite{lhg06,holg06} to approximate the hypergraph. 
    Each hyperedge of size $s$ is approximated by an $s$-clique.
    \item \textbf{Multi-layer perceptron (MLP)} treats each instance (hypernode) as an independent and identically distributed (i.i.d) instance. 
    In other words, $A=I$ in equation \ref{gcn}.
    We note that this baseline does not use the hypergraph structure to make predictions.
    \item \textbf{Multi-layer perceptron + explicit hypergraph Laplacian regularisation (MLP + HLR)}: regularises the MLP by training it with the loss given by $\mathcal{L} = \mathcal{L}_{0} + \lambda\mathcal{L}_{reg}$ and uses the hypergraph Laplacian with mediators for explicit Laplacian regularisation $L_{reg}$. We used $10\%$ of the test set used for all the above models for this baseline to get an optimal $\lambda$.
    \item \textbf{Confidence Interval-based method (CI) \cite{sslhg17}} uses a subgradient-based method \cite{sslhg17}.
    We note that this method has consistently been shown to be superior to the primal dual hybrid gradient (PDHG) of \cite{sslhg13} and also \cite{lhg06}. Hence, we did not use these other previous methods as baselines, and directly compared \method{} against CI.      
  \end{itemize}
 
The task for each dataset is to predict the topic to which a document belongs (multi-class classification).
Statistics are summarised in Table \ref{rel_view_data}.
For more details about datasets, please refer to the supplementary.
We trained all methods for $200$ epochs and used the same hyperparameters of a prior work \cite{gcniclr17}. 
We report the mean test error and standard deviation over $100$ different train-test splits.
We sampled sets of same sizes of labelled hypernodes from each class to have a balanced train split. 

\begin{table*}[t]
	\caption{  \label{rel_view_data}
		Real-world hypergraph datasets used in our work. Distribution of hyperedge sizes is not symmetric either side of the mean and has a strong positive skewness.}
	\centering
	\small
	\begin{tabular}{l|l|l|l|l|l}
		\hline
		& \textbf{DBLP} & \textbf{Pubmed} & \textbf{Cora} &\textbf{Cora} & \textbf{Citeseer}  \\
		& (co-authorship) & (co-citation)  & (co-authorship) & (co-citation) & (co-citation) \\
		\hline
		\# hypernodes, $|V|$ & $43413$ & $19717$ & $2708$   & $2708$ & $3312$ \\
		\# hyperedges, $|E|$ & $22535$  & $7963$ & $1072$ & $1579$ & $1079$ \\
		\textbf{avg.hyperedge size} & $\mathbf{4.7\pm6.1}$ & $\mathbf{4.3\pm5.7}$ & $\mathbf{4.2\pm4.1}$ & $\mathbf{3.0\pm1.1}$ & $\mathbf{3.2\pm2.0}$ \\		
		\# features, $d$ & $1425$ & $500$ & $1433$ & $1433$ & $3703$ \\
		\# classes, $q$ & $6$ & $3$ & $7$ & $7$ & $6$ \\
		label rate, $|V_L|/|V|$ & $0.040$ & $0.008$ & $0.052$ & $0.052$ & $0.042$ \\
		\hline
	\end{tabular}
\end{table*}

\begin{table*}[t]
	\caption{  \label{sslexp}
		Results of SSL experiments. We report mean test error $\pm$ standard deviation (lower is better) over $100$ train-test splits. Please refer to section \ref{sec:rel_exp} for details.}  
	\centering
	\resizebox{\textwidth}{!}{
	\begin{tabular}{llccccc}
		\hline\\
		\textbf{Data} &  \textbf{Method} & \textbf{DBLP} & \textbf{Pubmed} & \textbf{Cora} &\textbf{Cora} & \textbf{Citeseer}\\
		& & co-authorship & co-citation & co-authorship & co-citation & co-citation\\\\
		\hline\\
		$\mathbf{\mathcal{H}}$ & CI & $54.81\pm0.9$ & $52.96\pm0.8$ & $55.45\pm0.6$ & $64.40\pm0.8$ & $70.37\pm0.3$\\
		$\mathbf{X}$ & MLP & $37.77\pm2.0$  & $30.70\pm1.6$ & $41.25\pm1.9$ & $42.14\pm1.8$ & $41.12\pm1.7$ \\\\
		\hline\\
		$\mathbf{\mathcal{H}, X}$ & MLP + HLR & $30.42\pm2.1$  & $30.18\pm1.5$  & $34.87\pm1.8$ & $36.98\pm1.8$  & $37.75\pm1.6$\\
		$\mathbf{\mathcal{H}, X}$ & HGNN & $25.65\pm2.1$  & $29.41\pm1.5$ & $31.90\pm1.9$  & $\mathbf{32.41\pm1.8}$ & $\mathbf{37.40\pm1.6}$ \\\\
		\hline\\
		$\mathbf{\mathcal{H}, X}$ & 1-HyperGCN & $33.87\pm2.4$  & $30.08\pm1.5$ & $36.22\pm2.2$  & $34.45\pm2.1$  & $38.87\pm1.9$ \\
		$\mathbf{\mathcal{H}, X}$ & FastHyperGCN & $27.34\pm2.1$ & $29.48\pm1.6$  & $32.54\pm1.8$ & $\mathbf{32.43\pm1.8}$ & $\mathbf{37.42\pm1.7}$\\
		$\mathbf{\mathcal{H}, X}$ & HyperGCN & $\mathbf{24.09\pm2.0}$ & $\mathbf{25.56\pm1.6}$ & $\mathbf{30.08\pm1.8}$ & $\mathbf{32.37\pm1.7}$  & $\mathbf{37.35\pm1.6}$\\\\
		\hline
	\end{tabular}}
\end{table*}
\section{Analysis of results}
\label{sec:analysis}
The results on real-world datasets are shown in Table \ref{sslexp}. We now attempt to explain them.

\paragraph{Proposition 1:} Given a hypergraph $\mathcal{H}=(V, E)$ with $E\subseteq2^V-\cup_{v\in V}\{v\}$ and signals on the vertices $S:V\rightarrow{R}^d$, let, for each hyperedge $e\in E$, $(i_e, j_e):= \argmax_{i,j\in e}||S_i-S_j||_2$ and $K_e:=\{v\in e: v\neq i_e, v\neq j_e\}$. Define
\begin{itemize}
	\item $E_c:=\bigcup\limits_{e\in E}\Big\{\{u,v\}:u\in e, v\in e, u\neq v\Big\}$
	\item $w_c\Big(\{u,v\}\Big):=\sum\limits_{e\in E}\mathbbm{1}_{\{u,v\}\in E_c}\cdot\mathbbm{1}_{u\in e}\cdot\mathbbm{1}_{v\in e}\Big(\frac{2}{|e|\cdot(|e|-1)}\Big),\quad$	
	\item $E_m(S):=\bigcup\limits_{e\in E}\{i_e, j_e\}\quad\bigcup\quad\bigcup\limits_{e\in E, |e|\geq3}\Big\{\{u,v\}:u\in\{i_e, j_e\}, v\in K_e\big\}\Big\}$ 
	\item $w_m\Big(S, \{u,v\}\Big):=\sum\limits_{e\in E}\mathbbm{1}_{\{u,v\}\in E_m(S)}\cdot\mathbbm{1}_{u\in e}\cdot\mathbbm{1}_{v\in e}\Big(\frac{1}{2|e|-3}\Big),\quad$	
\end{itemize}
so that $G_c=(V, E_c, w_c)$ and $G_m(S)=(V, E_m(S), w_m(S))$ are the normalised clique exapnsion, i.e., graph of HGNN and mediator expansion, i.e., graph of \method{}/Fast\method{} respectively.
\textit{A sufficient condition for $G_c=G_m(S), \forall S$ is $\max\limits_{e\in E}|e| = 3$}.

\begin{table*}[t]
	\caption{  \label{preference}
		Results (lower is better) on sythetic data and a subset of DBLP showing that our methods are more effective for noisy hyperedges. $\eta$ is no. of hypernodes of one class divided by that of the other in noisy hyperedges. Best result is in bold and second best is underlined. Please see Section \ref{sec:analysis}.}  
	\centering
	\resizebox{\textwidth}{!}{
		\begin{tabular}{lccccccc}
			\hline
			 \textbf{Method} & $\eta=0.75$ & $\eta=0.70$ & $\eta=0.65$ &$\eta=0.60$ & $\eta=0.55$ & $\eta=0.50$ & sDBLP\\
			\hline
			 HGNN & $\mathbf{15.92\pm2.4}$  & $\mathbf{}$ $\mathbf{24.89\pm2.2}$  & $\mathbf{31.32\pm1.9}$ & $39.13\pm1.78$ & $42.23\pm1.9$ & $44.25\pm1.8$ & $45.27\pm2.4$ \\
			FastHyperGCN & $28.86\pm2.6$  & $31.56\pm2.7$ & $33.78\pm2.1$ & $\underline{33.89\pm2.0}$ & $\mathbf{34.56\pm2.2}$ & $\mathbf{35.65\pm2.1}$ & $\underline{41.79\pm2.8}$ \\
			HyperGCN & $\underline{22.44\pm2.0}$  & $\underline{29.33\pm2.2}$ & $\underline{33.41\pm1.9}$ & $\mathbf{33.67\pm1.9}$ & $\underline{35.05\pm2.0}$ & $\underline{37.89\pm1.9}$ & $\mathbf{41.64\pm2.6}$\\
	\end{tabular}}
\end{table*}

\paragraph{Proof:} Observe that we consider hypergraphs in which the size of each hyperedge is at least $2$. 
It follows from definitions that $|E_c|=\sum_{e\in E}\ ^{|e|}C_2$ and $|E_m|=\sum_{e\in E}\Big(2|e|-3\Big)$.
Clealy, a sufficient condition is when each hyperedge is approximated by the same subgraph in both the expansions.
In other words the condition is $\frac{|e|\cdot(|e|-1)}{2}=2|e|-3$ for each $e\in E$.
Solving the resulting quadratic eqution $x^2-5x+6=0$ gives us $(x-2)(x-3)=0$.
Hence, $|e|=2$ or $|e|=3$ for each $e\in E$. \hfill $\square$

\textbf{Comparable performance on Cora and Citeseer co-citation}
\hfill\break
We note that HGNN is the most competitive baseline. 
Also $S=X$ for Fast\method{} and $S=H\Theta$ for \method{}.
The proposition states that the graphs of HGNN,  Fast\method{}, and \method{} are the same irrespective of the signal values whenever the maximum size of a hyperedge is $3$.

This explains why the three methods have comparable accuracies for Cora co-citaion and Citeseer co-citiation hypergraphs.
The mean hyperedge sizes are close to $3$ (with comparitively lower deviations) as shown in Table \ref{rel_view_data}. 
Hence the graphs of the three methods are more or less the same.

\textbf{Superior performance on Pubmed, DBLP, and Cora co-authorship}
\hfill\break
We see that \method{} performs statistically significantly (p-value of Welch t-test is less than 0.0001) compared to HGNN on the other three datasets.
We believe this is due to large noisy hyperedges in real-world hypergraphs.
An author can write papers from different topics in a co-authorship network or a paper typically cites papers of different topics in co-citation networks.

Average sizes in Table \ref{rel_view_data} show the presence of large hyperedges (note the large standard deviations).
Clique expansion has edges on all pairs and hence potentially a larger number of hypernode pairs of different labels than the mediator graph of Figure \ref{fig:hmlap}, thus accumulating more noise.

\textbf{Preference of \method{} and Fast\method{} over HGNN}
\hfill\break
To further illustrate superiority over HGNN on noisy hyperedges, we conducted experiments on synthetic hypergraphs each consisting of $1000$ hypernodes, randomly sampled $500$ hyperedges, and $2$ classes with $500$ hypernodes in each class.
For each synthetic hypergraph, $100$ hyperedges (each of size $5$) were ``pure", i.e., all hypernodes were from the same class while the other $400$ hyperedges (each of size $20$) contained hypernodes from both classes.
The ratio, $\eta$, of hypernodes of one class to the other was varied from $0.75$ (less noisy) to $0.50$ (most noisy) in steps of $0.05$.

Table \ref{preference} shows the results on synthetic data.
We initialise the hypernode features to random Gaussian of $256$ dimensions.
We report mean error and deviation over $10$ different synthetically generated hypergraphs.
As we can see in the table for hyperedges with $\eta=0.75, 0.7$ (mostly pure), HGNN is the superior model.
However, as $\eta$ (noise) increases our methods begin to outperform HGNN.

\textbf{Subset of DBLP}: We also trained all three models on a subset of DBLP (we call it sDBLP) by removing all hyperedges of size $2$ and $3$.
The resulting hypergraph has around $8000$ hyperedges with an average size of $8.5\pm8.8$.
We report mean error over $10$ different train-test splits in
Table \ref{preference}.

\textbf{Conclusion}: From the above analysis, we conclude that our proposed methods (\method{} and Fast\method{}) should be preferred to HGNN for hypergraphs with large noisy hyperedges.
This is also the case on experiments in combinatorial optimisation (Table \ref{densetk}) which we discuss next.

\section{\method{} for combinatorial optimisation}
\label{sec:co_exp}

Inspired by the recent sucesses of deep graph models as learning-based approaches for NP-hard problems \cite{gtsgcn_nips18,tspgnn_aaai19,gcgnn_ijcai19,gatmco_kdd19}, we have used \method{} as a learning-based approach for the densest $k$-subhypergraph problem \cite{densestkhyper}.
NP-hard problems on hypergraphs have recently been highlighted as crucial for real-world network analysis \cite{hittingset_arxiv19,kcover_arxiv19}.
Our problem is, given a hypergraph $(V,E)$, to find a subset $W\subseteq V$ of $k$ hypernodes so as to maximise the number of hyperedges contained in $V$, i.e., we wish to maximise the density given by $|e\in E: e\subseteq W|$.

A greedy heuristic for the problem is to select the $k$ hypernodes of the maximum degree.
We call this ``MaxDegree".
Another greedy heuristic is to iteratively remove all hyperedges from the current (residual) hypergraph consisting of a hypernode of the minimum degree.
We repeat the procedure $n-k$ times and consider the density of the remaining $k$ hypernodes.
We call this ``RemoveMinDegree". 

\begin{table*}[t]
	\caption{  \label{densetk}
		Results on the densest $k$-subhypergraph problem. We report density (higher is better) of the set of vertices obtained by each of the proposed approaches for $k=\frac{3|V|}{4}$. See section \ref{sec:co_exp} for details.}  
	\centering
	\resizebox{\textwidth}{!}{
		\begin{tabular}{|l|c|c|c|c|c|c|}
			\hline
			\textbf{Dataset}$\rightarrow$ & \textbf{Synthetic} & \textbf{DBLP} & \textbf{Pubmed} & \textbf{Cora} &\textbf{Cora} & \textbf{Citeseer} \\
			Approach$\downarrow$ &  test set		&co-authorship & co-citation  & co-authorship & co-citation & co-citation \\
			\hline
			MaxDegree & $174\pm50$ & $4840$ & $1306$ & $194$ &  $544$& $507$ \\
			RemoveMinDegree & $147\pm48$ & $\mathbf{7714}$  & $\mathbf{7963}$ &$450$ &$1369$ & $843$  \\
			\hline\
			MLP & $174\pm56$ & $5580$ & $1206$ &$238$  & $550$ & $534$ \\
			MLP + HLR & $231\pm46$ & $5821$ & $3462$  &  $297$   & $952$  & $764$  \\
			HGNN & $337\pm49$ & $6274$ & $7865$ & $437$ & $\mathbf{1408}$ & $\mathbf{969}$ \\
			\hline
			$1$-\method{}  & $207\pm52$ & $5624$ & $1761$ & $251$  & $563$ & $509$  \\
			Fast\method{} & $352\pm45$ & $7342$ & $7893$ & $452$ &  $\mathbf{1419}$  & $\mathbf{969}$\\
			\method{} & $\mathbf{359\pm49}$ & $\mathbf{7720}$ & $\mathbf{7928}$ & $\mathbf{504}$ &  $\mathbf{1431}$  & $\mathbf{971}$\\
			\hline
			\textbf{\# hyperedges}, $|E|$ & $500$ &  $22535$  & $7963$ & $1072$ & $1579$ & $1079$ \\
			\hline
	\end{tabular}}
\end{table*}

\begin{figure*}[t]
	\centering
	\resizebox{\textwidth}{!}{\begin{tabular}{ccc}
			\centering
			\begin{minipage}{0.45\hsize}
				\centering
				\fbox{\includegraphics[scale=0.17]{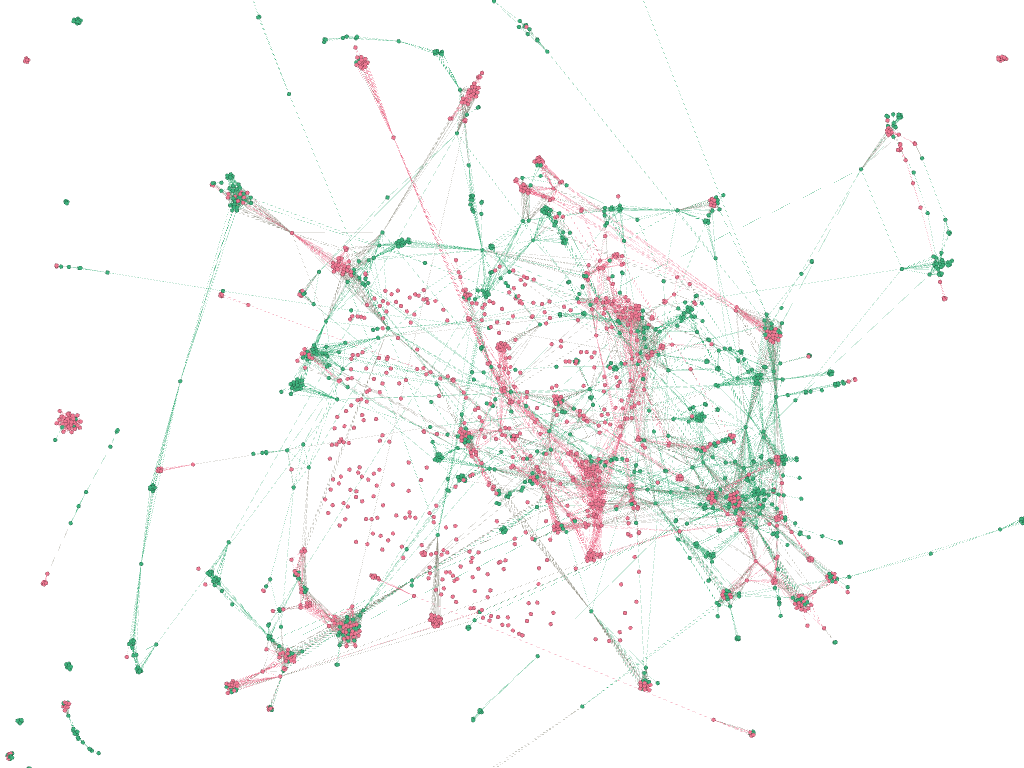}}\\
				\captionsetup{labelformat=empty}
				\subcaption{RemoveMinDegree}
				\label{}
			\end{minipage}
			\hspace{4mm}
			\begin{minipage}{0.45\hsize}
				\centering
				\fbox{\includegraphics[scale=0.17]{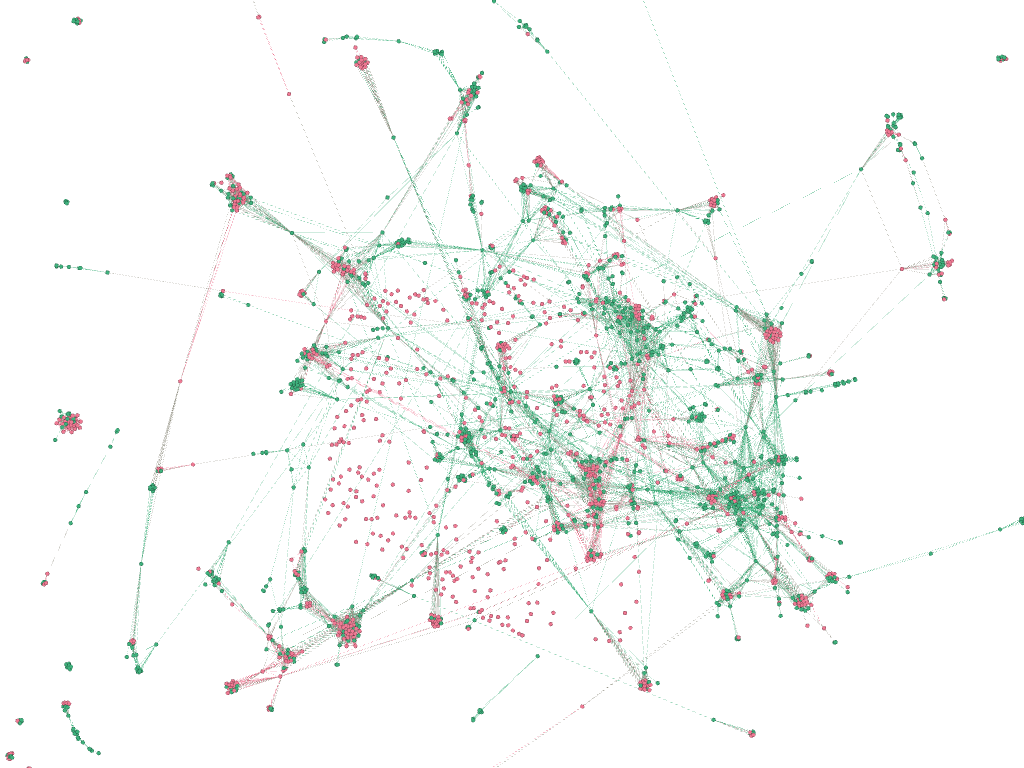}}\\
				\captionsetup{labelformat=empty}
				\subcaption{\method{}}
				\label{}
			\end{minipage}
	\end{tabular}}
	\caption{\label{coravis}Green / pink hypernodes denote those the algorithm labels as positive / negative respectively.}
\end{figure*}

\textbf{Experiments:} 
Table \ref{densetk} shows the results.
We trained all the learning-based models with a synthetically generated dataset.
More details on the approach and the synthetic data are in the supplementary.
As seen in Table \ref{densetk}, our proposed \method{} outperforms all the other approaches except for the pubmed dataset which contains a small number of vertices with large degrees and a large number of vertices with small degrees.
The RemoveMinDegree baseline is able to recover all the hyperedges here.

\textbf{Qualitative analysis}: 
Figure \ref{coravis} shows the visualisations given by RemoveMinDegree and \method{} on the Cora co-authorship hypergraph.
We used Gephi's Force Atlas to space out the vertices.
In general, a cluster of nearby vertices has multiple hyperedges connecting them.
Clusters of only green vertices indicate the method has likely included all vertices within the hyperedges induced by the cluster.
The figure of \method{} has more dense green clusters than that of RemoveMinDegree.
\section{Comparison of training time}
\label{train_time}
We compared the average training time of an epoch of Fast\method{} and HGNN in Table \ref{time_table}.
Both were run on a GeForce GTX 1080 Ti GPU machine.
We observe that Fast\method{} is faster than HGNN because it uses a linear number of edges for each hyperedge $e$ while HGNN uses quadratic.
Fast\method{} is also superior in terms of performance on hypergraphs with large noisy hyperedges. 

\section{Conclusion}
We have proposed \method{}, a new method of training GCN on hypergraph using tools from spectral theory of hypergraphs.
We have shown \method{}'s effectiveness in SSL and combinatorial optimisation.
Approaches that assign importance to nodes \cite{gat_iclr18,dpgcn_arxiv18,confgcn_aistats19} have improved results on SSL. 
\method{} may be augmented with such approaches for even more improved performance.
\title{Supplementary: Hypergraph convolutional network}

\section{Algorithms of our proposed methods}
The forward propagation of a $2$-layer graph convolutional network (GCN) \cite{gcniclr17} is 
\hfill\break
\[Z = \text{softmax}\Bigg(\bar{A}\ \ \text{ReLU}\bigg(\bar{A}X\Theta^{(1)}\bigg)\Theta^{(2)}\Bigg)\]
\hfill\break
$\text{ where }\ \ \bar{A} = \tilde{D}^{-\frac{1}{2}}\tilde{A}\tilde{D}^{-\frac{1}{2}},\quad \tilde{A} = A + I, \quad \text{and } \tilde{D}_{ii} = \sum_{j=1}^N\tilde{A}_{ij}$ and $D = \text{diag}(d_1,\cdots,d_N)$ is the diagonal degree matrix with elements $d_i=\sum_{j=1,j\neq i}^NA_{ji}$.
We provide algorithms for our three proposed methods:
\begin{itemize}
	\item \method{} - Algorithm \ref{hypergcnalgo}
	\item Fast\method{} - Algorithm \ref{fasthypergcnalgo}
	\item 1-\method{} - Algorithm \ref{1hypergcnalgo}
\end{itemize} 
\begin{algorithm}[H]
	\caption{\label{hypergcnalgo} Algorithm for HyperGCN}
	 \hspace*{\algorithmicindent} \textbf{Input}:  An attributed hypergraph $\mathcal{H}=(V,E,X)$, with attributes $X$, a set of labelled vertices $\mathcal{V}_L$\\ 
	  \hspace*{\algorithmicindent} \textbf{Output} All hypernodes in $V-\mathcal{V}_L$ labelled
	\begin{algorithmic}[1]
		\For{each epoch $\tau$ of training} 
		\For{layer $l=1,2$ of the network}  
		\State set $A^{(l)}_{vv}=1$ For all hypernodes $v\in V$ 
		\State let $\Theta=\Theta^{\tau}$ be the parameters For the current epoch
		\For{$e\in E$}
		\State $H\leftarrow$ hidden representation matrix of layer $l-1$
		\State $i_e,j_e:=\text{argmax}_{i,j \in e} ||H_i(\Theta^{(l)}) - H_j(\Theta^{(l)})||_2$
		\State $A^{(l)}_{i_e,j_e}=A^{(l)}_{j_e,i_e}=\frac{1}{2|e|-3}$
		\State $K_e:=\{k\in e:\ k\neq i_e, k\neq j_e\}$
		\For{$k\in K_e$}
		\State $A^{(l)}_{i_e,k}=A^{(l)}_{k,i_e}=\frac{1}{2|e|-3}$
		\State $A^{(l)}_{j_e,k}=A^{(l)}_{k,j_e}=\frac{1}{2|e|-3}$
		\EndFor
		\EndFor
		\EndFor
		\State $Z = \text{softmax}\Bigg(\bar{A}^{(2)}\ \ \text{ReLU}\bigg(\bar{A}^{(1)}X\Theta^{(1)}\bigg)\Theta^{(2)}\Bigg) $
		\State update parameters $\Theta^{\tau}$ to minimise cross entropy loss on the set of labelled hypernodes $\mathcal{V}_L$
		\EndFor
		\State label the hypernodes in $V-\mathcal{V}_L$ using $Z$ 
	\end{algorithmic}
\end{algorithm}

\begin{algorithm}[H]
	\caption{\label{fasthypergcnalgo} Algorithm for FastHyperGCN} 
	\hspace*{\algorithmicindent} \textbf{Input}:  An attributed hypergraph $\mathcal{H}=(V,E,X)$, with attributes $X$, a set of labelled vertices $\mathcal{V}_L$\\ 
	\hspace*{\algorithmicindent} \textbf{Output} All hypernodes in $V-\mathcal{V}_L$ labelled
	\begin{algorithmic}
		\State set $A_{vv}=1$ for all hypernodes $v\in V$ 
		\State $i_e,j_e:=\text{argmax}_{i,j \in e} ||X_i - X_j||_2$
		\For{$e\in E$}
		\State $A_{i_e,j_e}=A_{j_e,i_e}=\frac{1}{2|e|-3}$
		\State $K_e:=\{k\in e:\ k\neq i_e, k\neq j_e\}$
		\For{$k\in K_e$}
		\State $A_{i_e,k}=A_{k,i_e}=\frac{1}{2|e|-3}$
		\State $A_{j_e,k}=A_{k,j_e}=\frac{1}{2|e|-3}$
		\EndFor
		\EndFor
		\For{each epoch $\tau$ of training} 
		\State let $\Theta=\Theta^{\tau}$ be the parameters for the current epoch
		\State $Z = \text{softmax}\Bigg(\bar{A}\ \ \text{ReLU}\bigg(\bar{A}X\Theta^{(1)}\bigg)\Theta^{(2)}\Bigg) $
		\State update parameters $\Theta^{\tau}$ to minimise cross entropy loss on the set of labelled hypernodes $\mathcal{V}_L$
		\EndFor
		\State label the hypernodes in $V-\mathcal{V}_L$ using $Z$ 
	\end{algorithmic}
\end{algorithm}

\begin{algorithm}[H]
	\caption{\label{1hypergcnalgo} Algorithm for $1$-HyperGCN} 
	 \hspace*{\algorithmicindent} \textbf{Input}:  An attributed hypergraph $\mathcal{H}=(V,E,X)$, with attributes $X$, a set of labelled vertices $\mathcal{V}_L$\\ 
	\hspace*{\algorithmicindent} \textbf{Output} All hypernodes in $V-\mathcal{V}_L$ labelled
	\begin{algorithmic}
		\For{each epoch $\tau$ of training} 
		\For{layer $l=1,2$ of the network}  
		\State set $A^{(l)}_{vv}=1$ for all hypernodes $v\in V$ 
		\State let $\Theta=\Theta^{\tau}$ be the parameters for the current epoch
		\For{$e\in E$}
		\State $H\leftarrow$ hidden representation matrix of layer $l-1$
		\State $i_e,j_e:=\text{argmax}_{i,j \in e} ||H_i(\Theta^{(l)}) - H_j(\Theta^{(l)})||_2$
		\State $A^{(l)}_{i_e,j_e}=A^{(l)}_{j_e,i_e}=\frac{1}{|e|}$
		\EndFor
		\EndFor
		\State $Z = \text{softmax}\Bigg(\bar{A}^{(2)}\ \ \text{ReLU}\bigg(\bar{A}^{(1)}X\Theta^{(1)}\bigg)\Theta^{(2)}\Bigg) $
		\State update parameters $\Theta^{\tau}$ to minimise cross entropy loss on the set of labelled hypernodes $\mathcal{V}_L$
		\EndFor
		\State label the hypernodes in $V-\mathcal{V}_L$ using $Z$ 
	\end{algorithmic}
\end{algorithm}

\subsection{Time complexity}
Given an attributed hypergraph $(V, E, X)$, let $d$ be the number of initial features, $h$ be the number of hidden units, and $l$ be the number of labels.
Further, let $T$ be the total number of epochs of training.
Define
\[N:=\sum_{e\in E}|e|,\quad\quad\quad N_{m}:=\sum_{e\in E}\Big(2|e|-3\Big),\quad\quad\quad N_{c}:=\sum_{e\in E}\ ^{|e|}C_2\]
\begin{itemize}
	\item \method{} takes $O\bigg(T\Big(N+N_mh(d+c)\Big)\bigg)$ time
	\item 1-\method{} takes $O\bigg(TN\Big(1+h(d+c)\Big)\bigg)$ time
	\item Fast\method{} takes $O\bigg(TN_mh\big(d+c\big)\bigg)$ time
	\item HGNN takes $O\bigg(TN_ch\big(d+c\big)\bigg)$ time	
\end{itemize}

\section{\method{} for combinatorial optimisation}
\label{sec:sub_co_exp}
Inspired by the recent sucesses of deep graph models as learning-based approaches for NP-hard problems \cite{gtsgcn_nips18,tspgnn_aaai19,gcgnn_ijcai19,gatmco_kdd19},  we have used \method{} as a learning-based approach for the densest $k$-subhypergraph problem \cite{densestkhyper}, an NP-hard hypergraph problem.
The problem is given a hypergraph $(V,E)$, find a subset $W\subseteq V$ of $k$ hypernodes so as to maximise the number of hyperedges contained in (induced by) $V$ i.e. we intend to maximise the density given by 
\[|e\in E: e\subseteq W|\]
One natural greedy heuristic approach for the problem is to select the $k$ hypernodes of the maximum degree.
We call this approach ``MaxDegree".
Another greedy heuristic approach is to iteratively remove all the hyperedges from the current (residual) hypergraph containing a hypernode of the minimum degree.
We repeat the procedure $n-k$ times and consider the density of the remaining $k$ hypernodes.
We call this approach ``RemoveMinDegree". 

\subsection{Our approach}
A natural approach to the problem is to train \method{} to perform the labelling.
In other words, \method{} would take an input hypergraph $(V,E)$ as input and output a binary labelling of the hypernodes $v\in V$.
A natural output representation is a probability map in $[0,1]^{|V|}$ that indicates how likely each hypernode is to belong to $W$.

Let $\mathcal{D}=\{(V_i, E_i), l_i\}$ be a training set, where $(V_i, E_i)$ is an input hypergraph and $l_i\in\{0,1\}^{|V|\times1}$ is one of the optimal solutions for the NP-hard hypergraph problem.
The \method{} model learns its parameters $\Theta$ and is trained to predict $l_i$ given $(V_i,E_i)$.
During training we minimise the binary cross-entropy loss $L$ for each training sample $\{(V_i, E_i), l_i\}$
Additionally we generate $M$ different probability maps to minimise the hindsight loss i.e. $\sum_{i}\min_{m}L^{(m)}$
where $L^{(m)}$ is the cross-entropy loss corresponding to the $m$-th probability map.
Generating multiple probability maps has the advantage of generating diverse solutions \cite{gtsgcn_nips18}.

\subsection{Experiments: Training data}
To generate a sample $\{(V, E), l\}$ in the training set $\mathcal{D}$, we fix a vertex set $W$ of $k$ vertices chosen uniformly randomly.
We generate each hyperedge $e\in E$ such that $e\subseteq W$ with high probability $p$.
Note that $e\subseteq V-W$ with probability $1-p$.
We give the algorithm to generate a sample $\{(V, E), l\}$.

\begin{algorithm}[H]
	\caption{\label{syntheticalgo} Algorithm for generating a training sample} 
	\hspace*{\algorithmicindent} \textbf{Input}:   A hypergraph $(V, E)$ and a dense set of vertices $W$ $\mathcal{V}_L$\\ 
	\hspace*{\algorithmicindent} \textbf{Output} A hypergraph $(V, E)$ and a dense set of vertices $W$
	\begin{algorithmic}
		\State $|E|\leftarrow\frac{|V|}{2}$
		\State $W\leftarrow$ subset of $V$ of size $k$ chosen uniformly randomly
		\For {$i=1,2,\cdots,|E|$}
		\State $|e|\sim\{2,3,\cdots 10\}$ chosen uniformly randomly
		\State sample $e$ from $W$ with probability $p$
		\State sample $e$ from $V-W$ with probability $1-p$
		\EndFor
	\end{algorithmic}
\end{algorithm}

\subsection{Experiments: Results}
We generated $5000$ training samples with the number of hypernodes $|V|$ uniformly randomly chosen from $\{1000, 2000, \cdots, 5000\}$.
We fix $|E|=\frac{|V|}{2}$ as this is mostly the case for real-world hypergraphs.
Further we chose $e\in E$ such that $|e|$ is uniformly randomly chosen from $\{2,\cdots,10\}$ as this is also mostly the case for real-world hypergraphs.
We compared all our proposed approaches viz. $1$-\method{}, \method{}, and Fast\method{} against the baselines MLP, MLP+HLR and the state-of-the art HGNN.
We also compared against the greedy heuristics MaxDegree and RemoveMinDegree.
We train all the deep models using the same hyperparameters of \cite{gtsgcn_nips18} and report the results for $p=0.75$ and $k=\frac{3|V|}{4}$ in Table \ref{sub_densetk}.
We test all the models on a synthetically generated test set of hypergraphs with $1000$ vertices for each.
We also test the models on the five real-world hypergraphs used for SSL experiments.
As we can see in the table our proposed \method{} outperforms all the other approaches except for the pubmed dataset which contains a small number of vertices with large degrees and a large number of vertices with small degrees.
The RemoveMinDegree baseline is able to recover all the hyperedges in the pubmed dataset.
Moreover Fast\method{} is competitive with \method{} as the number of hypergraphs in the training data is large.

\subsection{Qualitative analysis}
\begin{figure*}[t]
	\centering
	\resizebox{\textwidth}{!}{\begin{tabular}{ccc}
			\centering
			\begin{minipage}{0.45\hsize}
				\centering
				\fbox{\includegraphics[scale=0.17]{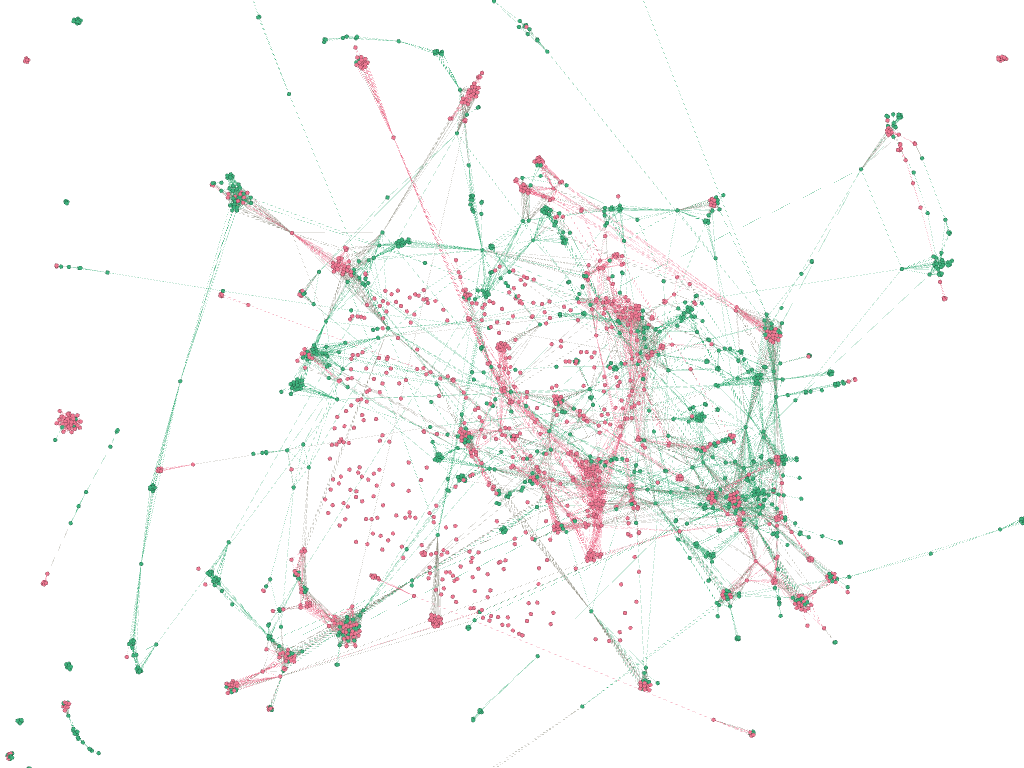}}\\
				\captionsetup{labelformat=empty}
				\subcaption{RemoveMinDegree}
				\label{}
			\end{minipage}
			\hspace{4mm}
			\begin{minipage}{0.45\hsize}
				\centering
				\fbox{\includegraphics[scale=0.17]{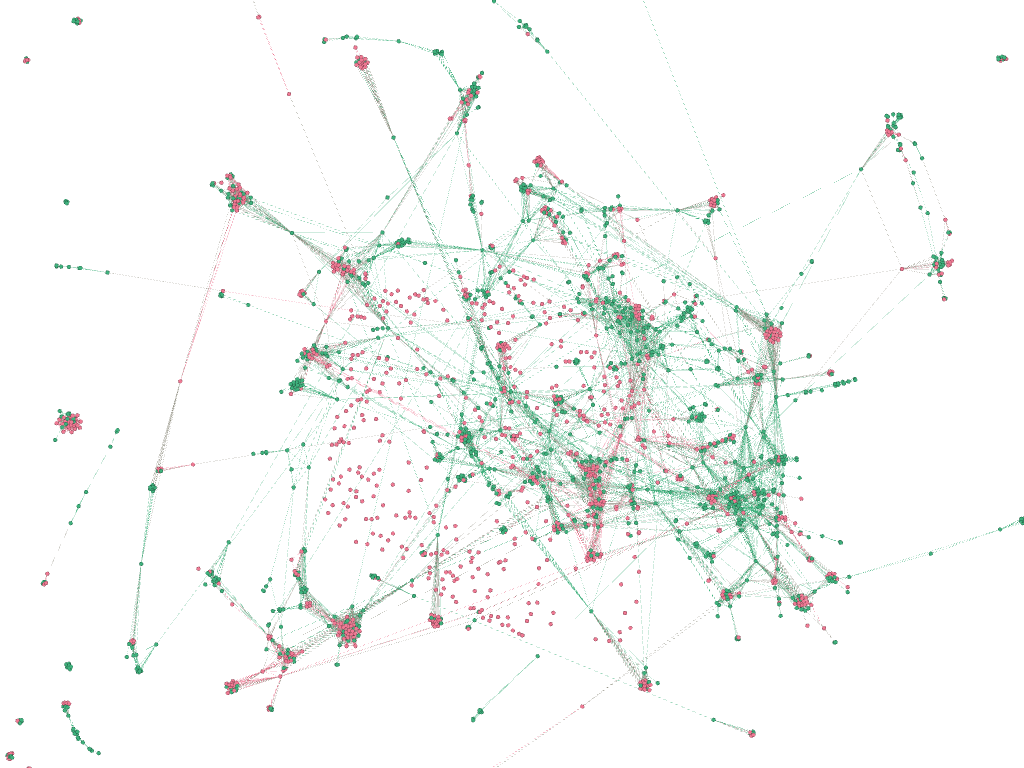}}\\
				\captionsetup{labelformat=empty}
				\subcaption{\method{}}
				\label{}
			\end{minipage}
	\end{tabular}}
	\caption{\label{sub_coravis}Green / pink hypernodes denote those the algorithm labels as positive / negative respectively.}
\end{figure*}

\begin{figure*}[t]
	\centering
	\begin{tabular}{ccc}
		\centering
		\begin{minipage}{0.45\hsize}
			\centering
			\fbox{\includegraphics[scale=0.17]{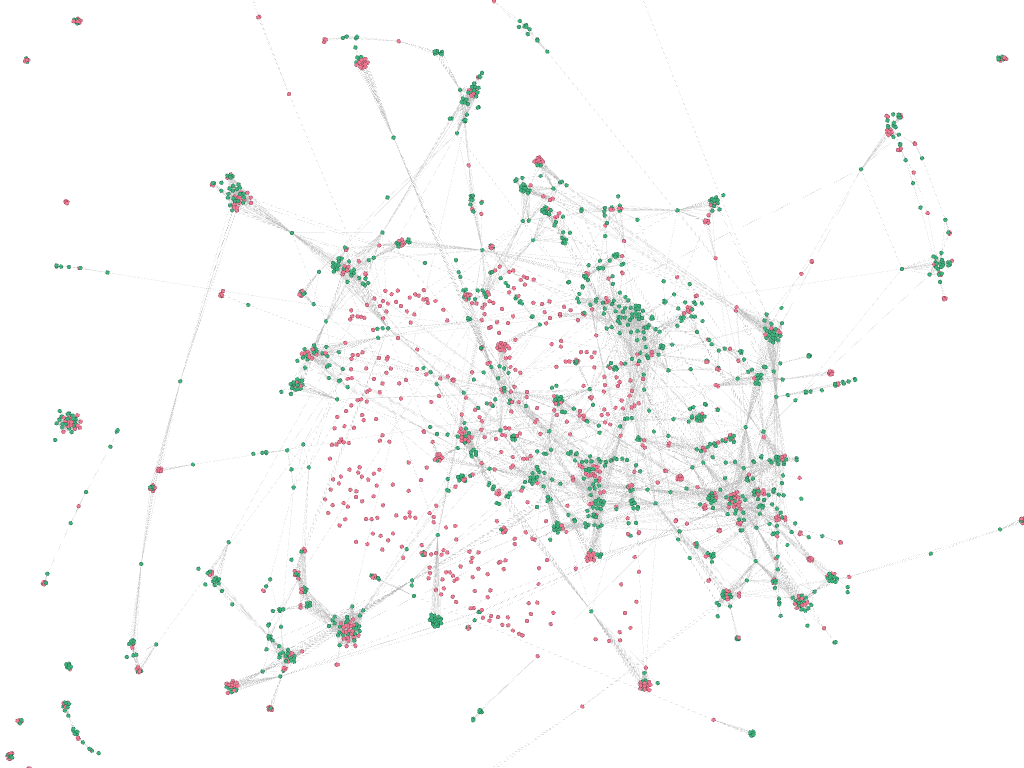}}\\
			\captionsetup{labelformat=empty}
			\subcaption{HGNN}
			\label{}
		\end{minipage}
		\hspace{4mm}
		\begin{minipage}{0.45\hsize}
			\centering
			\fbox{\includegraphics[scale=0.17]{mediators.png}}\\
			\captionsetup{labelformat=empty}
			\subcaption{\method{}}
			\label{}
		\end{minipage}
	\end{tabular}
	\caption{\label{coravishgnn}Green / pink hypernodes denote those the algorithm labels as positive / negative respectively.}
\end{figure*}

Figure \ref{sub_coravis} shows the visualisations given by RemoveMinDegree and \method{} on the Cora co-authorship hypergraph.
We used Gephi's Force Atlas to space out the vertices.
In general, a cluster of nearby vertices has multiple hyperedges connecting them.
Clusters of only green vertices indicate the method has likely included all vertices within the hyperedges induced by the cluster.
The figure of \method{} has more dense green clusters than that of RemoveMinDegree.
Figure \ref{coravishgnn} shows the results of HGNN vs. \method{}.

\begin{table*}[t]
	\caption{  \label{sub_densetk}
		Results on the densest $k$-subhypergraph problem. We report density (higher is better) of the set of vertices obtained by each of the proposed approaches for $k=\frac{3|V|}{4}$. See Section \ref{sec:sub_co_exp} for details.}  
	\centering
	\resizebox{\textwidth}{!}{
		\begin{tabular}{|l|c|c|c|c|c|c|}
			\hline
			\textbf{Dataset}$\rightarrow$ & \textbf{Synthetic} & \textbf{DBLP} & \textbf{Pubmed} & \textbf{Cora} &\textbf{Cora} & \textbf{Citeseer} \\
			Approach$\downarrow$ &  test set		&co-authorship & co-citation  & co-authorship & co-citation & co-citation \\
			\hline
			MaxDegree & $174\pm50$ & $4840$ & $1306$ & $194$ &  $544$& $507$ \\
			RemoveMinDegree & $147\pm48$ & $\mathbf{7714}$  & $\mathbf{7963}$ &$450$ &$1369$ & $843$  \\
			\hline\
			MLP & $174\pm56$ & $5580$ & $1206$ &$238$  & $550$ & $534$ \\
			MLP + HLR & $231\pm46$ & $5821$ & $3462$  &  $297$   & $952$  & $764$  \\
			HGNN & $337\pm49$ & $6274$ & $7865$ & $437$ & $\mathbf{1408}$ & $\mathbf{969}$ \\
			\hline
			$1$-\method{}  & $207\pm52$ & $5624$ & $1761$ & $251$  & $563$ & $509$  \\
			Fast\method{} & $352\pm45$ & $7342$ & $7893$ & $452$ &  $\mathbf{1419}$  & $\mathbf{969}$\\
			\method{} & $\mathbf{359\pm49}$ & $\mathbf{7720}$ & $\mathbf{7928}$ & $\mathbf{504}$ &  $\mathbf{1431}$  & $\mathbf{971}$\\
			\hline
			\textbf{\# hyperedges}, $|E|$ & $500$ &  $22535$  & $7963$ & $1072$ & $1579$ & $1079$ \\
			\hline
	\end{tabular}}
\end{table*}

\section{Sources of the real-world datasets}
\textbf{Co-authorship data}: All documents co-authored by an author are in one hyperedge. We used the author data\footnote{https://people.cs.umass.edu/~mccallum/data.html}to get the co-authorship hypergraph for cora.
We manually constructed the DBLP dataset from Arnetminer\footnote{https://aminer.org/lab-datasets/citation/DBLP-citation-Jan8.tar.bz}.

\textbf{Co-citation data}: All documents cited by a document are connected by a hyperedge. We used cora, citeseer, pubmed from \footnote{https://linqs.soe.ucsc.edu/data} for co-citation relationships.
We removed hyperedges which had exactly one hypernode as our focus in this work is on hyperedges with two or more hypernodes.
Each hypernode (document) is represented by bag-of-words features (feature matrix $X$).

\subsection{Construction of the DBLP dataset}
We downloaded the entire dblp data from \url{https://aminer.org/lab-datasets/citation/DBLP-citation-Jan8.tar.bz}.
The steps for constructing the dblp dataset used in the paper are as follows:
\begin{itemize}
  \item We defined a set of $6$ conference categories (classes for the SSL task) as ``algorithms", ``database", ``programming", ``datamining", ``intelligence", and ``vision"
  \item For a total of $4304$ venues in the entire dblp dataset we took papers from only a subset of venues from \url{https://en.wikipedia.org/wiki/List_of_computer_science_conferences} corresponding to the above $5$ conferences 
  \item From the venues of the above $5$ conference categories, we got $22535$ authors publishing at least two documents for a total of $43413$
  \item  We took the abstracts of all these $43413$ documents, constructed a dictionary of the most frequent words (words with frequency more than $100$) and this gave us a dictionary size of $1425$
\end{itemize}

\section{Experiments on datasets with categorical attributes}
\label{sec:cat_exp}

\begin{table*}[t]
  \caption{  \label{uci_data_sum}
Summary of the three UCI datasets used in the experiments in Section \ref{sec:cat_exp}}
  \centering
  \begin{tabular}{cccc}
    \hline
    property/dataset     & mushroom & covertype45  & covertype67      \\
    \hline
    number of hypernodes, $|V|$ & $8124$   & $12240$ & $37877$ \\
    number of hyperedges, $|E|$ & $112$ & $104$ & $125$ \\
    number of edges in clique expansion & $65,999,376$ & $143,008,092$ & $1,348,219,153$ \\
    number of classes, $q$ & $2$ & $2$ & $2$ \\
    \hline
  \end{tabular}
\end{table*}

\begin{figure*}
\includegraphics[width=\textwidth,height=\textheight,keepaspectratio]{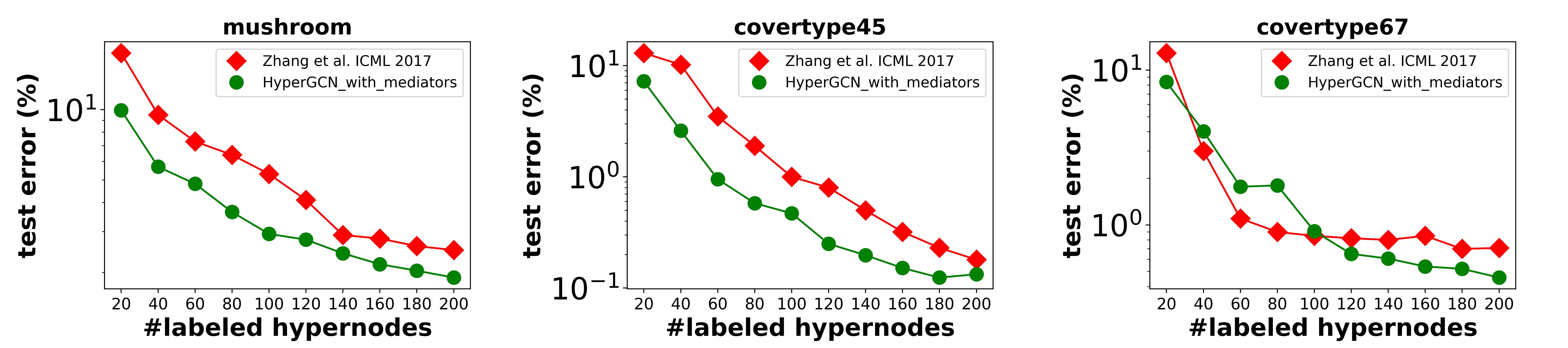}
\caption{\label{fig1}Test errors (lower is better) comparing \method{}\_with\_mediators with the non-neural baseline \cite{sslhg17} on the UCI datasets. \method{}\_with\_mediators offers superior performance. Comparing against GCN on Clique Expansion is unfair. Please see below  for details.}
\end{figure*}

We closely followed the experimental setup of the baseline model \cite{sslhg17}. We experimented on three different datasets viz., mushroom, covertype45, and covertype67 from the UCI machine learning repository \cite{uci17}. Properties of the datasets are summarised in Table \ref{uci_data_sum}.
The task for each of the three datasets is to predict one of two labels (binary classification) for each unlabelled instance (hypernode).
The datasets contain instances with categorical attributes.
To construct the hypergraph, we treat each attribute value as a hyperedge, i.e., all instances (hypernodes) with the same attribute value are contained in a hyperedge.
Because of this particular definition of a hyperedge clique expansion is destined to produce an almost fully connected graph and hence GCN on clique expansion will be unfair to compare against.
Having shown that \method{}  is superior to $1$-\method{} in the relational experiments, we compare only the former and the non-neural baseline \cite{sslhg17}.
We have called\method{}  as \method{}\_with\_mediators.
We used the incidence matrix (that encodes the hypergraph structure) as the data matrix $X$.
We trained \method{}\_with\_mediators for the full $200$ epochs and we used the same hyperparameters as in \cite{gcniclr17}.

As in \cite{sslhg17}, we performed $100$ trials for each $|V_L|$ and report the mean accuracy (averaged over the $100$ trials). The results are shown in Figure \ref{fig1}. We find that \method{}\_with\_mediators model generally does better than the baselines. We believe that this is because of the powerful feature extraction capability of \method{}\_with\_mediators.

\subsection{GCN on clique expansion}
We reiterate that clique expansion, i.e., HGNN \cite{hgnn_aaai19} for all the three datasets produce almost fuly connected graphs and hence clique expansion does not have any useful information.
So, GCN on clique expansion is unfair to compare against (HGNN does not learn any useful weights for classification because of the fully connected nature of the graph).

\subsection{Relevance of SSL}
The main reason for performing these experiments, as pointed out in the publicly accessible NIPS reviews\footnote{https://papers.nips.cc/paper/4914-the-total-variation-on-hypergraphs-learning-on-hypergraphs-revisited}   of the total variation on hypergraphs \cite{sslhg13}, is to show that the proposed method (the primal-dual hybrid gradient method in their case and the \method{}\_with\_mediators method in our case) has improved results on SSL, even if SSL is not very relevant in the first place.

We do not claim that SSL with \method{}\_with\_mediators is the best way to go about handling these categorical data but we do claim that, \textit{given this built hypergraph albeit from non-relational data}, it has superior results compared to the previous best non-neural hypergraph-based SSL method \cite{sslhg17} in the literature and that is why we have followed their experimental setup.

\section{Derivations}
We show how the graph convolutional network (GCN) \cite{gcniclr17} has its roots from the convolution theorem \cite{ctmallat99}.

\begin{table*}
	\caption{  \label{sup_pubmed_co_citation}
		Results on \textbf{\textit{Pubmed co-citation}} hypergraph. Mean test error $\pm$ standard deviation (lower is better) over $100$ trials  for different values of $|V_L|$. We randomly sampled the same number of labelled hypernodes from each class and hence we chose each $|V_L|$ to be divisible by $q$ with $\frac{|V_L|}{|V|}$ $0.2$ to $1\%$.}  
	\centering
	\resizebox{\textwidth}{!}{
	\begin{tabular}{llccccc}
		\hline\\
		\textbf{Available data} &  \textbf{Method} & $\mathbf{39}$ & $\mathbf{78}$ & $\mathbf{120}$ & $\mathbf{159}$ & $\mathbf{198}$\\
		& & $\mathbf{0.2\%}$ & $\mathbf{0.4\%}$ & $\mathbf{0.6\%}$ & $\mathbf{0.8\%}$ & $\mathbf{1\%}$\\\\
		\hline\\
		$\mathbf{\mathcal{H}}$ & CI  & $62.61\pm1.69$ & $58.53\pm1.25$ & $55.71\pm1.03$ & $52.96\pm0.79$  & $50.21\pm0.56$  \\
		$\mathbf{X}$ & MLP & $43.85\pm7.80$  & $35.17\pm4.92$ & $32.04\pm2.31$ & $30.70\pm1.61$ & $28.87\pm1.16$ \\\\
		\hline\\
		$\mathbf{\mathcal{H}, X}$ & MLP + HLR & $42.31\pm6.99$  & $33.69\pm4.49$ & $31.79\pm2.38$ & $30.18\pm1.54$ & $28.09\pm1.29$ \\
		$\mathbf{\mathcal{H}, X}$ & HGNN & $37.99\pm6.45$  & $33.01\pm4.25$ &  $31.14\pm2.23$ & $29.41\pm1.47$ & $26.96\pm1.35$ \\\\
		\hline\\
		$\mathbf{\mathcal{H}, X}$ & 1-HyperGCN & $43.62\pm7.18$  & $34.58\pm4.24$ & $31.88\pm2.78$ & $30.08\pm1.53$ & $28.90\pm1.29$\\
		$\mathbf{\mathcal{H}, X}$ & FastHyperGCN & $39.72\pm6.45$  & $32.67\pm3.91$ & $30.66\pm2.45$ & $29.48\pm1.60$ & $26.55\pm1.31$\\
		$\mathbf{\mathcal{H}, X}$ & HyperGCN &  $\mathbf{33.33\pm7.01}$ & $\mathbf{31.71\pm4.37}$ & $\mathbf{28.84\pm2.60}$ & $\mathbf{25.56\pm1.55}$ & $\mathbf{23.97\pm1.24}$
		\\\\
		\hline
	\end{tabular}}
\end{table*}

\subsection{Graph signal processing}
We now briefly review essential concepts of graph signal processing that are important in the construction of ChebNet and graph convolutional networks. 
 We need convolutions on graphs defined in the spectral domain.
 Similar to regular $1$-D or $2$-D signals, real-valued graph signals can be efficiently analysed via harmonic analysis and processed in the spectral domain \cite{shuman13}.
 To define spectral convolution, we note that the convolution theorem \cite{ctmallat99} generalises from classical discrete signal processing to take into account arbitrary graphs \cite{dspg13}.

Informally, the {\em convolution theorem\/} says the convolution of two signals in one domain (say time domain) equals point-wise multiplication of the signals in the other domain (frequency domain).
More formally, given a graph signal, $S:\mathcal{V}\rightarrow\mathbb{R}$, $S\in\mathbb{R}^N$, and a filter signal, $F:\mathcal{V}\rightarrow\mathbb{R}$, $F\in\mathbb{R}^N$, both of which are defined in the vertex domain (time domain), the convolution of the two signals, $C = S\star F$, satisfies
 \begin{equation}
 \label{ct}
 \hat{C} = \hat{S}\odot \hat{F}
 \end{equation}
where $\hat{S}$, $\hat{F}$, $\hat{C}$ are the graph signals in the spectral domain (frequency domain) corresponding, respectively, to $S$, $F$ and $S\star F$.

An essential operator for computing graph signals in the spectral domain is the symmetrically normalised graph Laplacian operator of $\mathcal{G}$, defined as
\begin{equation}
\label{graphLaplacian}
L = I - D^{-\frac{1}{2}}AD^{-\frac{1}{2}}
\end{equation} 
where $D = \text{diag}(d_1,\cdots,d_N)$ is the diagonal degree matrix with elements $d_i=\sum_{j=1,j\neq i}^NA_{ji}$.
As the above graph Laplacian operator, $L$, is a real symmetric and positive semidefinite matrix, it admits spectral eigen decomposition of the form $L=U\Lambda U^T$, where, $U=[u_1,\cdots,u_N]$ forms an orthonormal basis of eigenvectors and $\Lambda = \text{diag}(\lambda_1,\cdots,\lambda_N)$ is the diagonal matrix of the corresponding eigenvalues with $0=\lambda_1\leq\cdots\leq\lambda_N\leq2$.

The eigenvectors form a Fourier basis and the eigenvalues carry a notion of frequencies as in  classical Fourier analysis. 
The graph Fourier transform of a graph signal $S=(S_1,\cdots,S_N)\in\mathbb{R}^N$, is thus defined as $\hat{S} = U^TS$ and the inverse graph Fourier transform turns out to be $S=U\hat{S}$ , which is the same as, 
\begin{equation}
S_i = \sum_{j=1}^N\hat{S}(\lambda_j)u_j(i)\quad \text{ for }\ i\in \mathcal{V}=\{1,\cdots,N\}
\end{equation}
The convolution theorem generalised to graph signals ~\ref{ct} can thus be rewritten as $U^TC = \hat{S}\odot\hat{F}$.
It follows that $C = U(\hat{S}\odot\hat{F})$, which is the same as 
\begin{equation}
\label{graphconv}
C_i = \sum_{j=1}^N\hat{S}(\lambda_j)\hat{F}(\lambda_j)u_j(i) \quad \text{ for }\ i\in \mathcal{V}=\{1,\cdots,N\}
\end{equation}

\begin{table*}[t]
	\caption{  \label{sup_dblp_co_author}
		Results on \textbf{\textit{DBLP co-authorship}} hypergraph. Mean test error $\pm$ standard deviation (lower is better) over $100$ trials  for different values of $|V_L|$. We randomly sampled the same number of labelled hypernodes from each class and hence we chose each $|V_L|$ to be divisible by $q$ with $\frac{|V_L|}{|V|}$ $1$ to $5\%$.}  
	\centering
	\resizebox{\textwidth}{!}{
	\begin{tabular}{llccccc}
		\hline\\
		\textbf{Available data} &  \textbf{Method} & $\mathbf{438}$ & $\mathbf{870}$ & $\mathbf{1302}$ & $\mathbf{1740}$ & $\mathbf{2172}$\\
		& & $\mathbf{1\%}$ & $\mathbf{2\%}$ & $\mathbf{3\%}$ & $\mathbf{4\%}$ & $\mathbf{5\%}$\\\\
		\hline\\
		$\mathbf{\mathcal{H}}$ & CI  & $61.32\pm1.58$ & $59.39\pm1.37$ & $56.95\pm1.12$ & $54.81\pm0.94$ & $51.33\pm0.66$ \\
		$\mathbf{X}$ & MLP & $44.57\pm7.19$  & $42.23\pm4.88$ & $38.89\pm3.62$ & $37.77\pm2.02$ & $35.12\pm1.57$ \\\\
		\hline\\
		$\mathbf{\mathcal{H}, X}$ & MLP + HLR & $34.54\pm7.49$  & $33.50\pm4.17$ & $32.77\pm3.16$ & $30.42\pm2.07$ & $29.21\pm1.94$\\
		$\mathbf{\mathcal{H}, X}$ & HGNN & $30.62\pm8.02$  & $27.09\pm4.48$ & $26.18\pm3.29$ & $25.65\pm2.08$& $\mathbf{24.02\pm1.91}$ \\\\
		\hline\\
		$\mathbf{\mathcal{H}, X}$ & 1-HyperGCN & $40.17\pm6.99$  & $36.99\pm4.78$  & $34.44\pm3.43$ & $33.87\pm2.39$ & $32.11\pm1.96$ \\
		$\mathbf{\mathcal{H}, X}$ & FastHyperGCN & $34.03\pm7.59$  & $29.93\pm4.35$ & $28.57\pm3.13$ & $27.34\pm2.06$ & $25.23\pm1.84$\\
		$\mathbf{\mathcal{H}, X}$ & HyperGCN & $\mathbf{28.51\pm7.73}$  & $\mathbf{25.45\pm4.32}$ & $\mathbf{24.69\pm3.08}$ & $\mathbf{24.09\pm2.02}$ & $\mathbf{23.96\pm1.98}$\\\\
		\hline
	\end{tabular}}
\end{table*}

\subsection{ChebNet convolution}
We could use a non-parametric filter $\hat{F}(\lambda_j) = \theta_j$ \quad $\text{ for }\ j\in \{1,\cdots,N\}$ but there are two limitations: (i) they are not localised in space (ii) their learning complexity is $O(N)$.
The two limitations above contrast with with traditional CNNs where the filters are localised in space and the learning complexity is independent of the input size.
It is proposed by \cite{chebnet_nips16} to use a polynomial filter to overcome the limitations.
A polynomial filter is defined as: 
\begin{equation}
\label{polyfil}
\hat{F}(\lambda_j) = \sum_{k=0}^{K}w_k\lambda_j^k\quad\text{ for }\ j\in\{1,\cdots,N\}
\end{equation}
Using ~\ref{polyfil} in ~\ref{graphconv}, we get $C_i = \sum_{j=1}^N\hat{S}(\lambda_j)\bigg(\sum_{k=0}^{K}w_k\lambda_j^k\bigg)u_j(i) \quad \text{ for }\ i\in \mathcal{V}=\{1,\cdots,N\}$.
From the definition of an eigenvalue, we have $Lu_j=\lambda_ju_j$ and hence $L^ku_j=\lambda_j^ku_j$ for a positive integer $k$ and  $\text{ for }\ j\in\{1,\cdots,N\}$.
Therefore, 
\begin{equation} \label{chebnet}
\begin{split}
C_i & =  \sum_{j=1}^N\hat{S}(\lambda_j)\bigg(\sum_{k=0}^{K}w_kL_i^k\bigg)u_j(i) \\
 & = \bigg(\sum_{k=0}^{K}w_kL_i^k\bigg)\sum_{j=1}^N\hat{S}(\lambda_j)u_j(i) \\
 & = \bigg(\sum_{k=0}^{K}w_kL_i^k\bigg)S_i
\end{split}
\end{equation}
 Hence,
\begin{equation}
\label{chebnet_full}\ \  C=\bigg(\sum_{k=0}^{K}w_kL^k\bigg)S
\end{equation}
The graph convolution provided by Eq. ~\ref{chebnet_full} uses the monomial basis $1,x,\cdots,x^K$ to learn filter weights. 
Monomial bases are not optimal for training and not stable under perturbations because they do not form an orthogonal basis.
It is proposed by  \cite{chebnet_nips16} to use the orthogonal Chebyshev polynomials \cite{wgsgt11} (and hence the name ChebNet) to recursively compute the powers of the graph Laplacian.

A Chebyshev polynomial $T_k(x)$ of order $k$ can be computed recursively by the stable recurrence relation $T_k(x) = 2xT_{k-1}(x) - T_{k-2}(x)$ with $T_0 = 1$ and $T_1 = x$.
These polynomials form an orthogonal basis in $[-1,1]$.
Note that the eigenvalues of the symmetrically normalised graph Laplacian ~\ref{graphLaplacian} lie in the range $[0,2]$.
Through appropriate scaling of eigenvalues from $[0,2]$ to $[-1,1]$ i.e. $\tilde{\lambda}_j = \frac{2\lambda_j}{\lambda_N} - 1$ for $j=\{1,\cdots,N\}$, where $\lambda_N$ is the largest eigenvalue, the filter in ~\ref{polyfil} can be parametrised as the truncated expansion 
\begin{equation}
\label{chebexp}
\hat{F}(\lambda_j) = \sum_{k=0}^{K}w_kT_k(\tilde{\lambda}_j)\quad\text{ for }\ j\in\{1,\cdots,N\}
\end{equation}
From Eq. ~\ref{chebnet}, it follows that
\begin{equation}
\label{chebnet_simple}
C=\bigg(\sum_{k=0}^{K}w_kT_k(\tilde{L})\bigg)S\quad \text{where } \tilde{L} = \frac{2L}{\lambda_N} - I
\end{equation}

\begin{table*}[t]
	\caption{  \label{sup_cora_co_authorship}
		Results on \textbf{\textit{Cora co-authorship}} hypergraph. Mean test error $\pm$ standard deviation (lower is better) over $100$ trials  for different values of $|V_L|$. We randomly sampled the same number of labelled hypernodes from each class and hence we chose each $|V_L|$ to be divisible by $q$.}  
	\centering
	\small
	\begin{tabular}{llcccc}
		\hline\\
		\textbf{Available data} &  \textbf{Method} & $\mathbf{42}$ & $\mathbf{98}$ & $\mathbf{140}$ & $\mathbf{203}$\\\\
		\hline\\
		$\mathbf{\mathcal{H}}$ & CI & $67.72\pm 0.60$ & $58.55\pm 0.53$ & $55.45\pm 0.55$ & $51.44\pm 0.32$ \\
		$\mathbf{X}$ & MLP & $61.32\pm4.86$ & $47.69\pm2.36$ & $41.25\pm1.85$ & $37.76\pm1.32$\\\\
		\hline\\
		$\mathbf{\mathcal{H}, X}$ & MLP + HLR & $54.31\pm5.12$ & $41.06\pm2.53$ & $34.87\pm1.78$ & $32.21\pm1.43$ \\
		$\mathbf{\mathcal{H}, X}$ & HGNN & $45.23\pm 5.03$ & $34.08\pm 2.40$ & $31.90\pm 1.87$ & $\mathbf{28.92\pm 1.49}$ \\\\
		\hline\\
		$\mathbf{\mathcal{H}, X}$ & 1-HyperGCN & $50.26\pm4.78$ & $39.01\pm1.76$ & $36.22\pm2.21$ & $32.78\pm1.63$ \\
		$\mathbf{\mathcal{H}, X}$ & HyperGCN & $\mathbf{43.86\pm4.78}$ & $\mathbf{33.83\pm1.81}$ & $\mathbf{30.08\pm1.80}$ & $\mathbf{29.08\pm1.44}$
		\\\\
		\hline
	\end{tabular}
\end{table*}

\subsection{Graph convolutional network (GCN): first-order approximation of ChebNet}
The spectral convolution of ~\ref{chebnet_simple} is $K$-localised since it is a $K^{th}$-order polynomial in the Laplacian i.e. it depends only on nodes that are at most $K$ hops away.
\cite{gcniclr17} simplify ~\ref{chebnet_simple} to $K=1$ i.e. they use simple filters operating on $1$-hop neighbourhoods of the graph.
More formally,
\begin{equation}
\label{gcn_eq}
C = \bigg(w_0 + w_1\tilde{L}\bigg)S 
\end{equation}
and also,
\begin{equation}
\hat{F}(\lambda_j) = w_0 + w_1\tilde{\lambda_j}\quad\text{ for }\ j\in\{1,\cdots,N\}
\end{equation}
The main motivation here is that ~\ref{gcn_eq} is not limited to the explicit parameterisation given by the Chebyshev polynomials.
Intuitively such a model cannot overfit on local neighbourhood structures for graphs with very wide node degree distributions, common in real-world graph datasets such as citation networks, social networks, and knowledge graphs.

In this formulation, \cite{gcniclr17} further approximate $\lambda_N \approx 2$, as the neural network parameters can adapt to the change in scale during training.
To address overfitting issues and to minimise the number of matrix multiplications, they set $w_0=-w_1=\theta$.
~\ref{gcn_eq} now reduces to
\begin{equation}
\label{gcnsin}
C = \theta(I-\tilde{L})S = \theta(2I-L)S = \theta(I + D^{-\frac{1}{2}}AD^{-\frac{1}{2}})S
\end{equation}
The filter parameter $\theta$ is shared over the whole graph and successive application of a filter of this form $K$ times then effectively convolves the $K^{th}$-order neighbourhood of a node, where $K$ is the number of convolutional layers (depth) of the neural network model.
We note that the eigenvalues of $L$ are in $[0,2]$ and hence the eigenvalues of $2I-L = I + D^{-\frac{1}{2}}AD^{-\frac{1}{2}}$ are also in the range $[0,2]$.
Repeated application of this operator can therefore lead to numerical instabilities and exploding/vanishing gradients.
To alleviate this problem, a {\it renormalisation trick} can be used \cite{gcniclr17}: 
\begin{equation}
\label{renormtrick}
I + D^{-\frac{1}{2}}AD^{-\frac{1}{2}}\rightarrow \tilde{D}^{-\frac{1}{2}}\tilde{A}\tilde{D}^{-\frac{1}{2}}
\end{equation}
$\text{with } \tilde{A} = A + I \quad\text{and } \tilde{D}_{ii} = \sum_{j=1}^N\tilde{A}_{ij}$.
Generalising the above to $p$ signals contained in the matrix $X\in\mathbb{R}^{N\times p}$ (also called the data matrix), and $r$ filter maps contained in the matrix $\Theta\in\mathbb{R}^{p\times r}$, the output convolved signal matrix will be:
\begin{equation}
\label{graphconvolutionoutput}
\bar{A}X\Theta\quad\text{ where }\bar{A} = \tilde{D}^{-\frac{1}{2}}\tilde{A}\tilde{D}^{-\frac{1}{2}}
\end{equation}

\begin{table*}[t]
	\caption{  \label{sup_cora_co_citation}
		Results on \textbf{\textit{Cora co-citation}} hypergraph. Mean test error $\pm$ standard deviation (lower is better) over $100$ trials  for different values of $|V_L|$. We randomly sampled the same number of labelled hypernodes from each class and hence we chose each $|V_L|$ to be divisible by $q$.}  
	\centering
	\small
	\begin{tabular}{llcccc}
		\hline\\
		\textbf{Available data} &  \textbf{Method} & $\mathbf{42}$ & $\mathbf{98}$ & $\mathbf{140}$ & $\mathbf{203}$\\\\
		\hline\\
		$\mathbf{\mathcal{H}}$ & CI & $79.25\pm1.34$ & $70.89\pm1.94$ & $64.40\pm0.81$ & $62.22\pm0.72$ \\
		$\mathbf{X}$ & MLP & $63.31\pm5.23$ & $47.97\pm3.15$ & $42.14\pm1.78$ & $40.05\pm1.58$ \\\\
		\hline\\
		$\mathbf{\mathcal{H}, X}$ & MLP + HLR & $56.21\pm5.65$ & $43.32\pm3.27$ & $36.98\pm1.83$ & $33.88\pm1.46$  \\
		$\mathbf{\mathcal{H}, X}$ & HGNN & $50.39\pm5.42$ & $\mathbf{35.62\pm3.11}$ & $\mathbf{32.41\pm1.82}$ &  $\mathbf{29.78\pm1.55}$ \\\\
		\hline\\
		$\mathbf{\mathcal{H}, X}$ & 1-HyperGCN & $50.39\pm5.41$ & $38.01\pm3.12$ & $34.45\pm2.05$ & $31.67\pm1.57$\\
		$\mathbf{\mathcal{H}, X}$ & HyperGCN & $\mathbf{47.00\pm5.32}$ & $\mathbf{35.76\pm2.60}$ & $\mathbf{32.37\pm1.71}$ & $\mathbf{29.98\pm1.45}$
		\\\\
		\hline
	\end{tabular}
\end{table*}

\subsection{GCNs for graph-based semi-supervised node classification}
The GCN is conditioned on both the adjacency matrix $A$ (underlying graph structure) and the data matrix $X$ (input features). 
This allows us to relax certain assumptions typically made in graph-based SSL, for example, the cluster assumption \cite{ca02} made by the explicit Laplacian-based regularisation methods.
This setting is especially powerful in scenarios where the adjacency matrix contains information not present in the data (such as citation links between documents in a citation network or relations in a knowledge graph).
The forward model for a simple two-layer GCN takes the following simple form:
\begin{equation}
\label{gcn_supp}
Z = f_{GCN}(X,A) = \text{softmax}\Bigg(\bar{A}\ \ \text{ReLU}\bigg(\bar{A}X\Theta^{(0)}\bigg)\Theta^{(1)}\Bigg) 
\end{equation}
where $\Theta^{(0)}\in\mathbb{R}^{p\times h}$ is an input-to-hidden weight matrix for a hidden layer with $h$ hidden units and $\Theta^{(1)}\in\mathbb{R}^{h\times r}$ is a  hidden-to-output weight matrix.
The softmax activation function defined as $\text{softmax}(x_i) = \frac{\text{exp}(x_i)}{\sum_i\text{exp}(x_i)}$ is applied row-wise.

\paragraph{Training} For semi-supervised multi-class classification with $q$ classes, we then evaluate the cross-entropy error over all the set of labelled examples, $\mathcal{V}_L$:
\begin{equation}
\label{sslgcnloss}
\mathcal{L} = -\sum_{i\in \mathcal{V}_L}\sum_{j=1}^q Y_{ij}\ln Z_{ij} 
\end{equation}
The weights of the graph convolutional network, viz. $\Theta^{(0)}$ and $\Theta^{(1)}$, are trained using gradient descent. 
Using efficient sparse-dense matrix multiplications for computing, the computational complexity of evaluating Eq. ~\ref{gcn_supp} is $O(|\mathcal{E}|phr)$ which is linear in the number of graph edges.

\subsection{GCN as a special form of Laplacian smoothing}
GCNs can be interpreted as a special form of symmetric Laplacian smoothing \cite{co_self_gcn_aaai18}.
The Laplacian smoothing \cite{sp_fsd_95} on each of the $p$ input channels in the input feature matrix $X\in\mathbb{R}^{N\times p}$ is defined as:
\begin{equation}
\label{lapSmoothChannel}
\chi_{i} = (1-\gamma)x_i + \gamma\sum_{j}\frac{\tilde{A}_{ij}}{d_i}x_j
\quad i=1,\cdots,N
\end{equation} 
here $\tilde{A} = A + I$ and $d_i$ is the degree of node $i$. Equivalently the Laplacian smoothing can be written as $\chi = X - \gamma\tilde{D}^{-1}\tilde{L}X = (I - \gamma\tilde{D}^{-1}\tilde{L})X$ where $\tilde{L} = \tilde{D} - \tilde{A}$.
Here $0\leq\gamma\leq 1$ is a parameter which controls the weighting between the feature of the current vertex and those of its neighbours.
If we let $\gamma=1$, and replace the normalised Laplacian $\tilde{D}^{-1}\tilde{L}$ by the symmetrically normalised Laplacian $\tilde{D}^{-\frac{1}{2}}\tilde{L}\tilde{D}^{-\frac{1}{2}}$, then $\chi = (I - \tilde{D}^{-\frac{1}{2}}\tilde{L}\tilde{D}^{-\frac{1}{2}})X = \bar{A}X$, the same as in the expression ~\ref{graphconvolutionoutput}.

Hence the graph convolution in the GCN is a special form of (symmetric) Laplacian smoothing.
The Laplacian smoothing of Eq. ~\ref{lapSmoothChannel} computes the new features of a node as the weighted average of itself and its neighbours.
Since nodes in the same cluster tend to be densely connected, the smoothing makes their features similar, which makes the subsequent classification task much easier.
Repeated application of Laplacian smoothing many times over leads to over-smoothing - the node features within each connected component of the graph will converge to the same values \cite{co_self_gcn_aaai18}.

\section{Hyperparameters and more experiments on SSL}
Please see tables \ref{sup_pubmed_co_citation},
\ref{sup_dblp_co_author}, \ref{sup_cora_co_authorship}, \ref{sup_cora_co_citation}, and \ref{sup_citeseer_co_citation} for the results on all the real-world hypergraph datasets.

Following a prior work \cite{gcniclr17}, we used the following hyperparameters for all the models:
\begin{itemize}
	\item hidden layer size: $32$
	\item dropout rate: $0.5$
	\item learning rate: $0.01$
	\item weight decay: $0.0005$
	\item number of training epochs: $200$
	\item $\lambda$ for explicit Laplacian regularisation: $0.001$
\end{itemize}

\begin{table*}[t]
  \caption{  \label{sup_citeseer_co_citation}
Results on \textbf{\textit{Citeseer co-citation}} hypergraph. Mean test error $\pm$ standard deviation (lower is better) over $100$ trials  for different values of $|V_L|$. We randomly sampled the same number of labelled hypernodes from each class and hence we chose each $|V_L|$ to be divisible by $q$.}  
\centering
\small
  \begin{tabular}{llcccc}
    \hline\\
       \textbf{Available data} &  \textbf{Method} & $\mathbf{42}$ & $\mathbf{102}$ & $\mathbf{138}$ & $\mathbf{198}$\\\\
    \hline\\
$\mathbf{\mathcal{H}}$ & CI & $74.68\pm1.02$ & $71.90\pm0.82$ & $70.37\pm0.29$ & $68.84\pm0.24$ \\
$\mathbf{X}$ & MLP & $57.14\pm4.87$ & $45.80\pm2.43$ & $41.12\pm1.65$ & $39.09\pm1.32$  \\\\
\hline\\
$\mathbf{\mathcal{H}, X}$ & MLP + HLR & $53.21\pm4.65$ & $43.21\pm2.35$ & $37.75\pm1.59$ & $36.01\pm1.29$ \\
$\mathbf{\mathcal{H}, X}$ & HGNN & $\mathbf{50.75\pm4.73}$ & $\mathbf{39.67\pm2.21}$ & $\mathbf{37.40\pm1.61}$ & $\mathbf{35.20\pm1.35}$\\\\
\hline\\
$\mathbf{\mathcal{H}, X}$ & 1-HyperGCN & $52.48\pm5.43$ & $41.26\pm2.54$ & $38.87\pm1.93$ & $36.46\pm1.46$  \\
$\mathbf{\mathcal{H}, X}$ & HyperGCN & $\mathbf{50.39\pm5.13}$ & $\mathbf{39.68\pm2.27}$ & $\mathbf{37.35\pm1.62}$ & $\mathbf{35.40\pm1.22}$ 
 \\\\
    \hline
  \end{tabular}
\end{table*}

\bibliography{hypergcn}

\begin{thebibliography}{57}
\providecommand{\natexlab}[1]{#1}
\providecommand{\url}[1]{\texttt{#1}}
\expandafter\ifx\csname urlstyle\endcsname\relax
  \providecommand{\doi}[1]{doi: #1}\else
  \providecommand{\doi}{doi: \begingroup \urlstyle{rm}\Url}\fi

\bibitem[Agarwal et~al.(2006)Agarwal, Branson, and Belongie]{holg06}
Sameer Agarwal, Kristin Branson, and Serge Belongie.
\newblock Higher order learning with graphs.
\newblock In \emph{International Conference on Machine Learning (ICML)}, pages
  17--24, 2006.

\bibitem[Amburg et~al.(2019)Amburg, Kleinberg, and Benson]{hittingset_arxiv19}
Ilya Amburg, Jon Kleinberg, and Austin~R. Benson.
\newblock Planted hitting set recovery in hypergraphs.
\newblock \emph{CoRR, arXiv:1905.05839}, 2019.

\bibitem[Atwood and Towsley(2016)]{dcnn_nips16}
James Atwood and Don Towsley.
\newblock Diffusion-convolutional neural networks.
\newblock In \emph{Neural Information Processing Systems (NIPS)}, pages
  1993--2001. Curran Associates, Inc., 2016.

\bibitem[Battaglia et~al.(2018)Battaglia, Hamrick, Bapst, Sanchez{-}Gonzalez,
  Zambaldi, Malinowski, Tacchetti, Raposo, Santoro, Faulkner,
  G{\"{u}}l{\c{c}}ehre, Song, Ballard, Gilmer, Dahl, Vaswani, Allen, Nash,
  Langston, Dyer, Heess, Wierstra, Kohli, Botvinick, Vinyals, Li, and
  Pascanu]{gnet_arxiv18}
Peter~W. Battaglia, Jessica Hamrick, Victor Bapst, Alvaro Sanchez{-}Gonzalez,
  Vin{\'{\i}}cius~Flores Zambaldi, Mateusz Malinowski, Andrea Tacchetti, David
  Raposo, Adam Santoro, Ryan Faulkner, {\c{C}}aglar G{\"{u}}l{\c{c}}ehre,
  Francis Song, Andrew Ballard, Justin Gilmer, George Dahl, Ashish Vaswani,
  Kelsey Allen, Charles Nash, Victoria Langston, Chris Dyer, Nicolas Heess,
  Daan Wierstra, Pushmeet Kohli, Matthew Botvinick, Oriol Vinyals, Yujia Li,
  and Razvan Pascanu.
\newblock Relational inductive biases, deep learning, and graph networks.
\newblock \emph{arXiv:1806.01261}, 2018.

\bibitem[Bronstein et~al.(2017)Bronstein, Bruna, LeCun, Szlam, and
  Vandergheynst]{gdl17}
Michael Bronstein, Joan Bruna, Yann LeCun, Arthur Szlam, and Pierre
  Vandergheynst.
\newblock Geometric deep learning: Beyond euclidean data.
\newblock \emph{{IEEE} Signal Process.}, 34:\penalty0 18--42, 2017.

\bibitem[Bul\`{o} and Pelillo(2009)]{gthg09}
Samuel~R. Bul\`{o} and Marcello Pelillo.
\newblock A game-theoretic approach to hypergraph clustering.
\newblock In \emph{Advances in Neural Information Processing Systems (NIPS)
  22}, pages 1571--1579. Curran Associates, Inc., 2009.

\bibitem[Chan and Liang(2018)]{laplacian_mediators_cocoon18}
T.{-}H.~Hubert Chan and Zhibin Liang.
\newblock Generalizing the hypergraph laplacian via a diffusion process with
  mediators.
\newblock In \emph{Computing and Combinatorics - 24th International Conference,
  (COCOON)}, pages 441--453, 2018.

\bibitem[Chan et~al.(2018)Chan, Louis, Tang, and Zhang]{hgl_18}
T.{-}H.~Hubert Chan, Anand Louis, Zhihao~Gavin Tang, and Chenzi Zhang.
\newblock Spectral properties of hypergraph laplacian and approximation
  algorithms.
\newblock \emph{J. {ACM}}, 65\penalty0 (3):\penalty0 15:1--15:48, 2018.

\bibitem[Chapelle et~al.(2003)Chapelle, Weston, and Sch\"{o}lkopf]{ca02}
Olivier Chapelle, Jason Weston, and Bernhard Sch\"{o}lkopf.
\newblock Cluster kernels for semi-supervised learning.
\newblock In \emph{Neural Information Processing Systems (NIPS)}, pages
  601--608. MIT, 2003.

\bibitem[Chapelle et~al.(2010)Chapelle, Scholkopf, and Zien]{csz_ssl10}
Olivier Chapelle, Bernhard Scholkopf, and Alexander Zien.
\newblock \emph{Semi-Supervised Learning}.
\newblock The MIT Press, 2010.

\bibitem[Chen et~al.(2019)Chen, Wei, Wang, and Guo]{mlgcn_cvpr19}
Zhao-Min Chen, Xiu-Shen Wei, Peng Wang, and Yanwen Guo.
\newblock Multi-label image recognition with graph convolutional networks.
\newblock In \emph{The IEEE Conference on Computer Vision and Pattern
  Recognition (CVPR)}, 2019.

\bibitem[Chien et~al.(2019)Chien, Zhou, and Li]{hgal_aistats19}
I~(Eli) Chien, Huozhi Zhou, and Pan Li.
\newblock $hs^2$: Active learning over hypergraphs with pointwise and pairwise
  queries.
\newblock In \emph{International Conference on Artificial Intelligence and
  Statistics (AISTATS)}, pages 2466--2475, 2019.

\bibitem[Chlamt{\'{a}}c et~al.(2018)Chlamt{\'{a}}c, Dinitz, Konrad, Kortsarz,
  and Rabanca]{densestkhyper}
Eden Chlamt{\'{a}}c, Michael Dinitz, Christian Konrad, Guy Kortsarz, and George
  Rabanca.
\newblock The densest k-subhypergraph problem.
\newblock \emph{{SIAM} J. Discrete Math.}, pages 1458--1477, 2018.

\bibitem[Defferrard et~al.(2016)Defferrard, Bresson, and
  Vandergheynst]{chebnet_nips16}
Micha\"{e}l Defferrard, Xavier Bresson, and Pierre Vandergheynst.
\newblock Convolutional neural networks on graphs with fast localized spectral
  filtering.
\newblock In \emph{Advances in Neural Information Processing Systems (NIPS)},
  pages 3844--3852. Curran Associates, Inc., 2016.

\bibitem[Dheeru and Karra~Taniskidou(2017)]{uci17}
Dua Dheeru and Efi Karra~Taniskidou.
\newblock {UCI} machine learning repository, 2017.
\newblock URL \url{http://archive.ics.uci.edu/ml}.

\bibitem[Feng et~al.(2018)Feng, He, Liu, Nie, and Chua]{poh18}
Fuli Feng, Xiangnan He, Yiqun Liu, Liqiang Nie, and Tat-Seng Chua.
\newblock Learning on partial-order hypergraphs.
\newblock In \emph{Proceedings of the 2018 World Wide Web Conference (WWW)},
  pages 1523--1532, 2018.

\bibitem[Feng et~al.(2019)Feng, You, Zhang, Ji, and Gao]{hgnn_aaai19}
Yifan Feng, Haoxuan You, Zizhao Zhang, Rongrong Ji, and Yue Gao.
\newblock Hypergraph neural networks.
\newblock In \emph{Proceedings of the Thirty-Third Conference on Association
  for the Advancement of Artificial Intelligence (AAAI)}, 2019.

\bibitem[Gilmer et~al.(2017)Gilmer, Schoenholz, Riley, Vinyals, and
  Dahl]{mpnn_icml17}
Justin Gilmer, Samuel~S. Schoenholz, Patrick~F. Riley, Oriol Vinyals, and
  George~E. Dahl.
\newblock Neural message passing for quantum chemistry.
\newblock In \emph{Proceedings of the 34th International Conference on Machine
  Learning (ICML)}, pages 1263--1272, 2017.

\bibitem[Gong et~al.(2019)Gong, Zhu, Duan, Liu, Guan, Sun, Ou, and
  Zhu]{gatmco_kdd19}
Yu~Gong, Yu~Zhu, Lu~Duan, Qingwen Liu, Ziyu Guan, Fei Sun, Wenwu Ou, and
  Kenny~Q. Zhu.
\newblock Exact-k recommendation via maximal clique optimization.
\newblock In \emph{Proceedings of the 25th {ACM} {SIGKDD} International
  Conference on Knowledge Discovery {\&} Data Mining (KDD)}, 2019.

\bibitem[Hamilton et~al.(2017)Hamilton, Ying, and Leskovec]{grl17}
William~L. Hamilton, Rex Ying, and Jure Leskovec.
\newblock Representation learning on graphs: Methods and applications.
\newblock \emph{{IEEE} Data Eng. Bull.}, 40\penalty0 (3):\penalty0 52--74,
  2017.

\bibitem[Hammond et~al.(2011)Hammond, Vandergheynst, and Gribonval]{wgsgt11}
David~K. Hammond, Pierre Vandergheynst, and R{\'e}mi Gribonval.
\newblock {Wavelets on graphs via spectral graph theory}.
\newblock \emph{{Applied and Computational Harmonic Analysis}}, 2011.

\bibitem[Hein et~al.(2013)Hein, Setzer, Jost, and Rangapuram]{sslhg13}
Matthias Hein, Simon Setzer, Leonardo Jost, and Syama~Sundar Rangapuram.
\newblock The total variation on hypergraphs - learning on hypergraphs
  revisited.
\newblock In \emph{Advances in Neural Information Processing Systems (NIPS)
  26}, pages 2427--2435. Curran Associates, Inc., 2013.

\bibitem[Kipf and Welling(2017)]{gcniclr17}
Thomas~N Kipf and Max Welling.
\newblock Semi-supervised classification with graph convolutional networks.
\newblock In \emph{International Conference on Learning Representations
  (ICLR)}, 2017.

\bibitem[Kolda and Bader(2009)]{td09}
Tamara~G. Kolda and Brett~W. Bader.
\newblock Tensor decompositions and applications.
\newblock \emph{SIAM Rev.}, 51\penalty0 (3):\penalty0 455--500, 2009.

\bibitem[Lemos et~al.(2019)Lemos, Prates, Avelar, and Lamb]{gcgnn_ijcai19}
Henrique Lemos, Marcelo Prates, Pedro Avelar, and Luis Lamb.
\newblock Graph colouring meets deep learning: Effective graph neural network
  models for combinatorial problems.
\newblock In \emph{Proceedings of the 28th International Joint Conference on
  Artificial Intelligence, (IJCAI)}, 2019.

\bibitem[Li and Milenkovic(2018{\natexlab{a}})]{dsfm_neurips18}
Pan Li and Olgica Milenkovic.
\newblock Revisiting decomposable submodular function minimization with
  incidence relations.
\newblock In \emph{Advances in Neural Information Processing Systems (NeurIPS)
  31}, pages 2237--2247. Curran Associates, Inc., 2018{\natexlab{a}}.

\bibitem[Li and Milenkovic(2018{\natexlab{b}})]{shg_icml18}
Pan Li and Olgica Milenkovic.
\newblock Submodular hypergraphs: p-laplacians, {C}heeger inequalities and
  spectral clustering.
\newblock In \emph{Proceedings of the 35th International Conference on Machine
  Learning (ICML)}, pages 3014--3023, 2018{\natexlab{b}}.

\bibitem[Li et~al.(2018{\natexlab{a}})Li, He, and Milenkovic]{qdsfm_neurips18}
Pan Li, Niao He, and Olgica Milenkovic.
\newblock Quadratic decomposable submodular function minimization.
\newblock In \emph{Advances in Neural Information Processing Systems (NeurIPS)
  31}, pages 1054--1064. Curran Associates, Inc., 2018{\natexlab{a}}.

\bibitem[Li et~al.(2018{\natexlab{b}})Li, Han, and Wu]{co_self_gcn_aaai18}
Qimai Li, Zhichao Han, and Xiao{-}Ming Wu.
\newblock Deeper insights into graph convolutional networks for semi-supervised
  learning.
\newblock In \emph{Proceedings of the Thirty-Second Conference on Association
  for the Advancement of Artificial Intelligence (AAAI)}, pages 3538--3545,
  2018{\natexlab{b}}.

\bibitem[Li et~al.(2018{\natexlab{c}})Li, Chen, and Koltun]{gtsgcn_nips18}
Zhuwen Li, Qifeng Chen, and Vladlen Koltun.
\newblock Combinatorial optimization with graph convolutional networks and
  guided tree search.
\newblock In \emph{Advances in Neural Information Processing Systems (NIPS)
  31}, pages 537--546. Curran Associates, Inc., 2018{\natexlab{c}}.

\bibitem[Louis(2015)]{hg_stoc15}
Anand Louis.
\newblock Hypergraph markov operators, eigenvalues and approximation
  algorithms.
\newblock In \emph{Proceedings of the Forty-Seventh Annual {ACM} on Symposium
  on Theory of Computing, (STOC)}, pages 713--722, 2015.

\bibitem[Mallat(1999)]{ctmallat99}
Stphane Mallat.
\newblock \emph{A Wavelet Tour of Signal Processing}.
\newblock Academic Press, 1999.

\bibitem[Marcheggiani and Titov(2017)]{gcn_srl_emnlp17}
Diego Marcheggiani and Ivan Titov.
\newblock Encoding sentences with graph convolutional networks for semantic
  role labeling.
\newblock In \emph{Proceedings of the 2017 Conference on Empirical Methods in
  Natural Language Processing (EMNLP)}, pages 1506--1515, 2017.

\bibitem[Monti et~al.(2018)Monti, Shchur, Bojchevski, Litany, G{\"{u}}nnemann,
  and Bronstein]{dpgcn_arxiv18}
Federico Monti, Oleksandr Shchur, Aleksandar Bojchevski, Or~Litany, Stephan
  G{\"{u}}nnemann, and Michael Bronstein.
\newblock Dual-primal graph convolutional networks.
\newblock \emph{abs/1806.00770}, 2018.

\bibitem[Nguyen et~al.(2019)Nguyen, Thai, Thai, Vu, and Dinh]{kcover_arxiv19}
Hung Nguyen, Phuc Thai, My~Thai, Tam Vu, and Thang Dinh.
\newblock Approximate k-cover in hypergraphs: Efficient algorithms, and
  applications.
\newblock \emph{CoRR, arXiv:1901.07928}, 2019.

\bibitem[Norcliffe-Brown et~al.(2018)Norcliffe-Brown, Vafeias, and
  Parisot]{vqagcn_nips18}
Will Norcliffe-Brown, Efstathios Vafeias, and Sarah Parisot.
\newblock Learning conditioned graph structures for interpretable visual
  question answering.
\newblock In \emph{Advances in Neural Information Processing Systems (NeurIPS)
  31}, pages 8344--8353. Curran Associates, Inc., 2018.

\bibitem[Prates et~al.(2019)Prates, Avelar, Lemos, Lamb, and
  Vardi]{tspgnn_aaai19}
Marcelo O.~R. Prates, Pedro H.~C. Avelar, Henrique Lemos, Luis Lamb, and Moshe
  Vardi.
\newblock Learning to solve np-complete problems - a graph neural network for
  the decision tsp.
\newblock In \emph{Proceedings of the Thirty-Third Conference on Association
  for the Advancement of Artificial Intelligence (AAAI)}, 2019.

\bibitem[Sandryhaila and Moura(2013)]{dspg13}
Aliaksei Sandryhaila and Jos{\'{e}} M.~F. Moura.
\newblock Discrete signal processing on graphs.
\newblock \emph{{IEEE} Trans. Signal Processing}, 2013.

\bibitem[Satchidanand et~al.(2015)Satchidanand, Ananthapadmanaban, and
  Ravindran]{hg_ijcai15}
Sai~Nageswar Satchidanand, Harini Ananthapadmanaban, and Balaraman Ravindran.
\newblock Extended discriminative random walk: {A} hypergraph approach to
  multi-view multi-relational transductive learning.
\newblock In \emph{Proceedings of the Twenty-Fourth International Joint
  Conference on Artificial Intelligence, (IJCAI)}, pages 3791--3797, 2015.

\bibitem[Shashua et~al.(2006)Shashua, Zass, and Hazan]{teeccv06}
Amnon Shashua, Ron Zass, and Tamir Hazan.
\newblock Multi-way clustering using super-symmetric non-negative tensor
  factorization.
\newblock In \emph{Proceedings of the 9th European Conference on Computer
  Vision (ECCV)}, pages 595--608, 2006.

\bibitem[Shuman et~al.(2013)Shuman, Narang, Frossard, Ortega, and
  Vandergheynst]{shuman13}
David~I. Shuman, Sunil~K. Narang, Pascal Frossard, Antonio Ortega, and Pierre
  Vandergheynst.
\newblock The emerging field of signal processing on graphs: Extending
  high-dimensional data analysis to networks and other irregular domains.
\newblock \emph{{IEEE} Signal Process. Mag.}, 30\penalty0 (3), 2013.

\bibitem[Subramanya and Talukdar(2014)]{ssl_ppt_14}
Amarnag Subramanya and Partha~Pratim Talukdar.
\newblock \emph{Graph-Based Semi-Supervised Learning}.
\newblock Morgan \& Claypool Publishers, 2014.

\bibitem[Sun et~al.(2018)Sun, Wang, Yu, and Li]{aagsurvey_arxiv18}
Lichao Sun, Ji~Wang, Philip~S. Yu, and Bo~Li.
\newblock Adversarial attack and defense on graph data: A survey.
\newblock \emph{CoRR, arXiv:1812.10528}, 2018.

\bibitem[Taubin(1995)]{sp_fsd_95}
Gabriel Taubin.
\newblock A signal processing approach to fair surface design.
\newblock In \emph{Proceedings of the 22Nd Annual Conference on Computer
  Graphics and Interactive Techniques}, 1995.

\bibitem[Vashishth et~al.(2019{\natexlab{a}})Vashishth, Bhandari, Yadav, Rai,
  Bhattacharyya, and Talukdar]{wordgcn_acl19}
Shikhar Vashishth, Manik Bhandari, Prateek Yadav, Piyush Rai, Chiranjib
  Bhattacharyya, and Partha Talukdar.
\newblock Incorporating syntactic and semantic information in word embeddings
  using graph convolutional networks.
\newblock In \emph{Proceedings of the 57th Annual Meeting of the Association
  for Computational Linguistics (ACL)}, 2019{\natexlab{a}}.

\bibitem[Vashishth et~al.(2019{\natexlab{b}})Vashishth, Yadav, Bhandari, and
  Talukdar]{confgcn_aistats19}
Shikhar Vashishth, Prateek Yadav, Manik Bhandari, and Partha Talukdar.
\newblock Confidence-based graph convolutional networks for semi-supervised
  learning.
\newblock In \emph{International Conference on Artificial Intelligence and
  Statistics (AISTATS)}, 2019{\natexlab{b}}.

\bibitem[Veli{\v{c}}kovi{\'{c}} et~al.(2018)Veli{\v{c}}kovi{\'{c}}, Cucurull,
  Casanova, Romero, Li{\`{o}}, and Bengio]{gat_iclr18}
Petar Veli{\v{c}}kovi{\'{c}}, Guillem Cucurull, Arantxa Casanova, Adriana
  Romero, Pietro Li{\`{o}}, and Yoshua Bengio.
\newblock Graph attention networks.
\newblock In \emph{International Conference on Learning Representations
  (ICLR)}, 2018.

\bibitem[Wang et~al.(2018)Wang, Ye, and Gupta]{zero_shot_gcn_cvpr18}
Xiaolong Wang, Yufei Ye, and Abhinav Gupta.
\newblock Zero-shot recognition via semantic embeddings and knowledge graphs.
\newblock In \emph{The IEEE Conference on Computer Vision and Pattern
  Recognition (CVPR)}, pages 6857--6866, 2018.

\bibitem[Weston et~al.(2008)Weston, Ratle, and Collobert]{dlvssl_icml08}
Jason Weston, Fr{\'e}d{\'e}ric Ratle, and Ronan Collobert.
\newblock Deep learning via semi-supervised embedding.
\newblock In \emph{Proceedings of the 25th International Conference on Machine
  Learning (ICML)}, pages 1168--1175, 2008.

\bibitem[Wu et~al.(2019)Wu, Pan, Chen, Long, Zhang, and Yu]{gnnsurvey_arxiv19}
Zonghan Wu, Shirui Pan, Fengwen Chen, Guodong Long, Chengqi Zhang, and
  Philip~S. Yu.
\newblock A comprehensive survey on graph neural networks.
\newblock \emph{CoRR, arXiv:1901.00596}, 2019.

\bibitem[Yao et~al.(2019)Yao, Mao, and Luo]{textgcn_aaai19}
Liang Yao, Chengsheng Mao, and Yuan Luo.
\newblock Graph convolutional networks for text classification.
\newblock In \emph{Proceedings of the Thirty-Third Conference on Association
  for the Advancement of Artificial Intelligence (AAAI)}, 2019.

\bibitem[Zhang et~al.(2017)Zhang, Hu, Tang, and Chan]{sslhg17}
Chenzi Zhang, Shuguang Hu, Zhihao~Gavin Tang, and T-H.~Hubert Chan.
\newblock Re-revisiting learning on hypergraphs: Confidence interval and
  subgradient method.
\newblock In \emph{Proceedings of 34th International Conference on Machine
  Learning (ICML)}, pages 4026--4034, 2017.

\bibitem[Zhang et~al.(2018)Zhang, Cui, and Zhu]{dlgsurvey_arxiv18}
Ziwei Zhang, Peng Cui, and Wenwu Zhu.
\newblock Deep learning on graphs: A survey.
\newblock \emph{CoRR, arXiv:1812.04202}, 2018.

\bibitem[Zhou et~al.(2003)Zhou, Bousquet, Lal, Weston, and
  Sch\"{o}lkopf]{sslintronips03}
Dengyong Zhou, Olivier Bousquet, Thomas~Navin Lal, Jason Weston, and Bernhard
  Sch\"{o}lkopf.
\newblock Learning with local and global consistency.
\newblock In \emph{NIPS}, 2003.

\bibitem[Zhou et~al.(2007)Zhou, Huang, and Sch\"{o}lkopf]{lhg06}
Denny Zhou, Jiayuan Huang, and Bernhard Sch\"{o}lkopf.
\newblock Learning with hypergraphs: Clustering, classification, and embedding.
\newblock In \emph{Advances in Neural Information Processing Systems (NIPS)
  19}, pages 1601--1608. MIT Press, 2007.

\bibitem[Zhu et~al.(2003)Zhu, Ghahramani, and Lafferty]{sslintroicml03}
Xiaojin Zhu, Zoubin Ghahramani, and John Lafferty.
\newblock Semi-supervised learning using gaussian fields and harmonic
  functions.
\newblock In \emph{ICML}, 2003.

\bibitem[Zhu et~al.(2009)Zhu, Goldberg, Brachman, and Dietterich]{sslintro09}
Xiaojin Zhu, Andrew~B. Goldberg, Ronald Brachman, and Thomas Dietterich.
\newblock \emph{Introduction to Semi-Supervised Learning}.
\newblock Morgan and Claypool Publishers, 2009.

\end{thebibliography}
\bibliographystyle{plainnat}
\end{document}